\documentclass[11pt]{article}

\usepackage{acl}

\usepackage{times}
\usepackage{latexsym}
\usepackage[T1]{fontenc}
\usepackage[utf8]{inputenc}
\usepackage{microtype}
\usepackage{inconsolata}

\usepackage{graphicx}
\usepackage{subcaption}
\usepackage{booktabs}
\usepackage{amsmath}
\usepackage{amssymb}
\usepackage{multirow}
\usepackage{array}
\usepackage{enumitem}
\usepackage{float}
\usepackage{placeins}
\usepackage{xcolor}
\usepackage{url}

\setlist[itemize]{leftmargin=1.4em}
\title{Beyond Endpoint Gains: A Weight-Delta Audit of Medical Specialization}

\author{
\textbf{Praphul Singh}\textsuperscript{\(\dagger,\ddagger,*\)}
\qquad
\textbf{Shanu Kumar}\textsuperscript{\(\ddagger,\mathsection\)}
\qquad
\textbf{Akshat Agarwal}\textsuperscript{\(\dagger,\ddagger\)}
\\[0.55em]
{\normalfont
\textsuperscript{\(\dagger\)}Oracle Health AI
\quad\textperiodcentered\quad
\textsuperscript{\(\ddagger\)}IIT Kanpur
\quad\textperiodcentered\quad
\textsuperscript{\(\mathsection\)}MBZUAI}
\\[0.35em]
{\small\normalfont
\textsuperscript{*}\textbf{Correspondence:}
\href{mailto:praphul.singh@oracle.com}{\texttt{praphul.singh@oracle.com}}}
}

\begin{document}
\maketitle

\begin{abstract}
Specialist language models are usually understood through endpoint gains: the
generalist scores lower, the specialist scores higher, and the difference is
treated as evidence of specialization. This leaves the released update itself
largely unexamined. We propose a paired weight-delta path audit and apply it to
two public, aligned generalist-to-medical-specialist checkpoint pairs:
Gemma-3-4B-IT\(\rightarrow\)MedGemma-4B-IT and
Qwen2.5-7B-Instruct\(\rightarrow\)HuatuoGPT-o1-7B. In both pairs, the full
decoder-side update strongly reconstructs measured medical benchmark movement
(0.974 and 1.183 endpoint-normalized retention), making each decoder delta an
appropriate substrate for the audit. Yet the movement is not cleanly localized.
MLP is the strongest broad component family in both pairs, but mixed
off-domain movements, 10-seed matched controls, and endpoint-anchored rollbacks
prevent a unique coarse-family explanation. The audit therefore separates
update-level reconstruction from component-level explanation. Its claims concern
text-only multiple-choice benchmark movement, not clinical validation, repair,
or circuit-level mechanism.
\end{abstract}

\section{Introduction}

Domain-specialized language models are usually presented through endpoint
comparisons. A base checkpoint is evaluated, a specialist checkpoint is
evaluated, and the score difference is taken as the evidence of specialization.
For medicine, this endpoint view is especially common: public exam-style
benchmarks provide a convenient way to summarize whether a medical model improves
over its generalist ancestor
\citep{Singhal_2023,Singhal_2025,nori2023capabilitiesgpt4medicalchallenge,sellergren2026medgemmatechnicalreport}.
But endpoint scores leave a basic question unanswered: what does the released
update itself carry?

This question matters because specialization is not a single scalar. The same
checkpoint update that improves a target-domain benchmark may also introduce
off-domain gains, off-domain regressions, and broad internal changes that are
not visible from the final score alone. Conversely, a component family that
appears useful in isolation may simply contain a large fraction of the update,
rather than uniquely explaining the behavior. Endpoint evaluation tells us that
two models differ; it does not audit whether the released update is broad or
localized, target-specific or mixed, or genuinely component-structured.

We study this problem using two public aligned generalist-to-medical-specialist
pairs: Gemma-3-4B-IT\(\rightarrow\)MedGemma-4B-IT
\citep{gemmateam2025gemma3technicalreport,sellergren2026medgemmatechnicalreport}
and Qwen2.5-7B-Instruct\(\rightarrow\)HuatuoGPT-o1-7B
\citep{yang2024qwen25technicalreport,chen2024huatuogpto1}. The former is our
primary detailed audit; the latter tests whether its central pattern survives a
change in architecture and checkpoint lineage. Because corresponding decoder
tensors are aligned within each pair, the released checkpoint difference can be
treated as an auditable path through weight space. We use medicine because it is
consequential and widely benchmarked, but our object is measured benchmark
movement along released updates, not clinical reliability.

\begin{figure*}[t]
    \centering
    \includegraphics[width=0.86\textwidth]{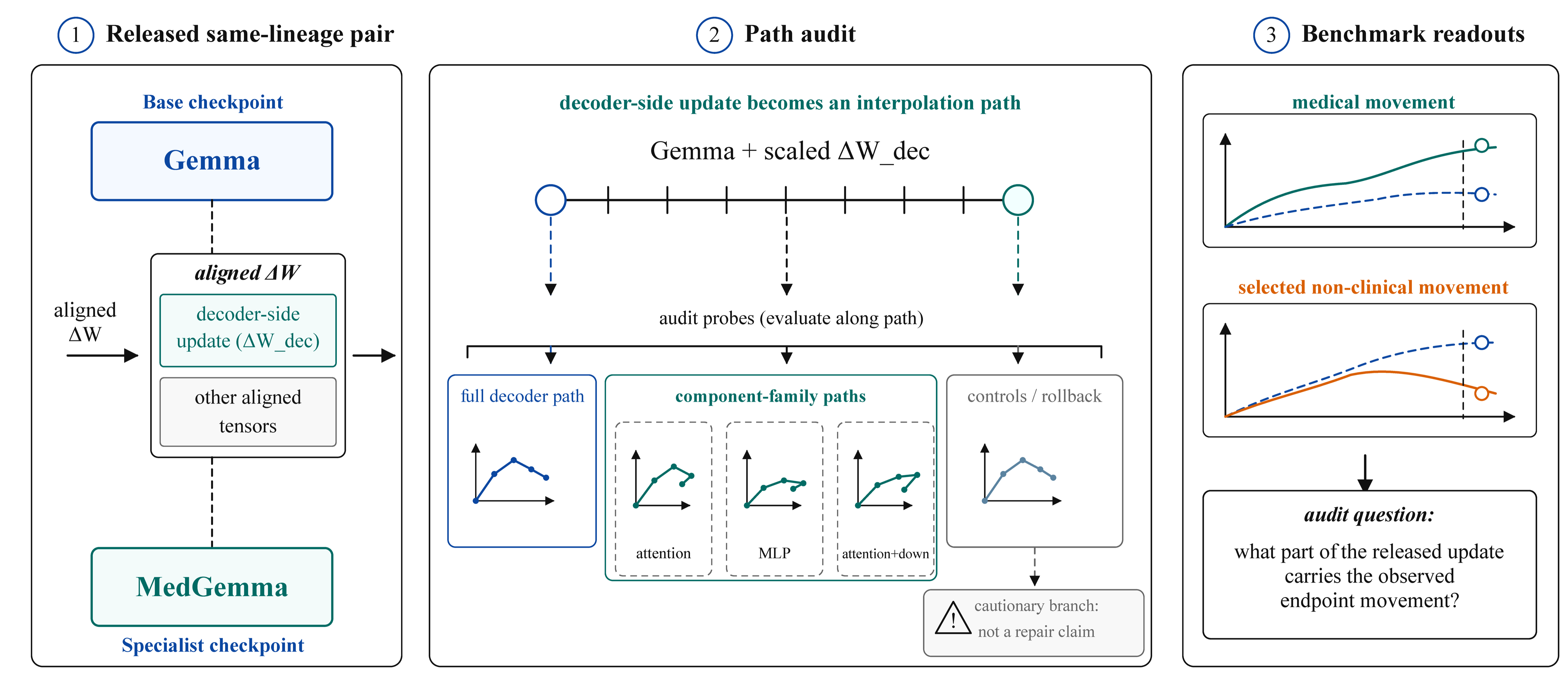}
    \caption{Paired checkpoint audit protocol, illustrated with the primary
    Gemma\(\rightarrow\)MedGemma pair. For each aligned release, we first test
    whether the full decoder-side path reconstructs measured benchmark movement,
    then whether component-family interpretations survive matched controls and
    endpoint rollback.}
    \label{fig:delta-path-overview}
\end{figure*}

This path view turns an endpoint comparison into a sequence of audit questions.
First, what changed in the released checkpoint? A descriptive \(\Delta W\)
screen shows that the primary update is broad and structured across decoder
layers and projection families, rather than a tiny localized edit. Second, does
the observed decoder update actually reconstruct the measured medical gain?
Evaluating the full decoder-side path shows that it does in both pairs.
Third, what else moves along the same path? Selected non-clinical regressions
and selected non-clinical gains coexist, while aggregate non-clinical accuracy
is nearly flat, so the update is not well summarized as broad forgetting.

The final question is whether this movement has a simple component-level
explanation. We test scoped component-family paths, matched random controls, and
endpoint-anchored rollbacks. These probes show partial sufficiency but not clean
localization. The MLP-family path is strongest among the tested scoped buckets,
but it is also large; matched random controls can approach or exceed structured
buckets; and rolling back individual families from the specialist endpoint
trades medical retention against selected-regression recovery without revealing
a clean knob. Thus the main component result is negative: the medical benchmark
movement is path-reproducible, but not uniquely localized to a tested coarse
component family.

This framing differs from task-vector transfer, model merging, model editing,
and circuit discovery. Those lines of work often ask how to compose, optimize,
repair, or mechanistically explain model behavior. We ask a narrower audit
question: given an aligned specialist release, what benchmark movement is carried
by its observed update, and how far do simple component explanations survive
controls? The evidential order matters: endpoint movement motivates full-update
reconstruction; reconstruction enables component partial-sufficiency tests;
matched controls test distinctiveness; and rollback tests intervention utility.
None of these stages alone supplies a circuit-level mechanism.

In this paper, we formulate a paired weight-delta path audit and evaluate it on
two aligned medical-specialist releases. The audit uses a medical
multiple-choice composite, a broader non-clinical suite, selected regression and
gain diagnostics, and decoder-side path sweeps. It contributes a reusable
protocol, a two-pair reconstruction result showing that measured medical movement
is carried by each released decoder update, and a negative localization lesson:
component claims must survive size/update-mass controls and endpoint rollback
before becoming explanations.

\section{Related Work}

\paragraph{Medical language models and endpoint evaluation.}
Medical language models are commonly evaluated with public exam-style
benchmarks, including MedQA and medical MMLU subsets
\citep{Jin_2021,hendrycks2021measuring}. Specialist systems and medical model
releases report gains on these benchmarks through prompting, post-training, or
domain-specific checkpoints
\citep{Singhal_2023,Singhal_2025,nori2023capabilitiesgpt4medicalchallenge,chen2023meditron70bscalingmedicalpretraining,sellergren2026medgemmatechnicalreport}.
Such results are useful behavioral summaries, but they are endpoint summaries:
they do not show how the update that produced the gain is organized, nor do they
establish reasoning quality or readiness for clinical deployment
\citep{Zhou_2025,ren2026medicalreasoninglargelanguage}. Our work is complementary: we
use an auditable medical multiple-choice composite as a controlled readout for
studying the released specialization update itself.

\paragraph{Mechanistic localization and intervention analysis.}
Interpretability work studies where model behaviors are implemented, including
feed-forward memories, knowledge neurons, attention heads, circuits, and sparse
features
\citep{petroni-etal-2019-language,roberts-etal-2020-much,geva-etal-2021-transformer,dai-etal-2022-knowledge,geva-etal-2023-dissecting,elhage2021mathematical,wang2023interpretability,cunningham2023sparseautoencodershighlyinterpretable,bricken2023monosemanticity,marks2025sparse}.
Causal mediation, causal tracing, attribution patching, and related
interventions provide finer-grained evidence for particular prompts or
mechanisms
\citep{NEURIPS2020_92650b2e,NEURIPS2022_6f1d43d5,NEURIPS2023_34e1dbe9,syed-etal-2024-attribution,hanna2024faithfaithfulnessgoingcircuit}.
Our component paths operate at a coarser granularity: they ask whether broad
subsets of an observed update reproduce benchmark movement, not whether a
minimal circuit or neuron set implements medical specialization.
Prior work gives reasons to expect important MLP contributions, but a large
family can look sufficient simply because it contains much of the update. We
therefore treat an MLP result as a hypothesis to test against size- and
energy-matched subsets, not as localization by itself.

\paragraph{Weight-space editing, interpolation, and merging.}
Weight-space methods modify or combine checkpoints through factual editing,
model averaging, task arithmetic, and interference-aware merging
\citep{NEURIPS2022_6f1d43d5,meng2023massediting,pmlr-v162-wortsman22a,NEURIPS2022_70c26937,ilharco2023editing,NEURIPS2023_1644c9af,song2026modelmergingeralarge}.
Activation- and representation-space interventions provide another control
surface for steering behavior
\citep{turner2024steeringlanguagemodelsactivation,zou2025representationengineeringtopdownapproach,rimsky-etal-2024-steering},
and paired-model weight patching is especially close in spirit
\citep{sun2026weightpatchingsourcelevelmechanistic}. These methods often treat deltas as objects to
transfer, compose, optimize, edit, or causally explain. Our use of the delta is
more modest: we treat the released same-lineage specialist update as evidence to
audit, asking which benchmark movements appear along it and how far component
explanations survive controls. In short, prior weight-delta work usually asks
how to change or explain a model; we ask what a released specialist update can
and cannot account for under benchmark readouts and negative controls.

\paragraph{Specialization tradeoffs and forgetting.}
Specialization and instruction tuning can improve target-domain behavior, while
their effects on broader capabilities require separate evaluation
\citep{bommasani2022opportunitiesrisksfoundationmodels,sellergren2026medgemmatechnicalreport,NEURIPS2022_b1efde53}. Continual-learning
work often studies forgetting across a known training sequence; in our setting,
only two released endpoints are observed. We therefore treat selected
non-clinical regressions and gains as endpoint-conditioned diagnostics of this
checkpoint pair, not as a complete estimate of forgetting or capability
transfer.

\section{Setup, Data, and Metrics}
\label{sec:setup-data}

\subsection{Paired Checkpoints}

Our primary pair is Gemma-3-4B-IT and its medical-specialized descendant,
MedGemma-4B-IT \citep{sellergren2026medgemmatechnicalreport}. We replicate the
audit on Qwen2.5-7B-Instruct and HuatuoGPT-o1-7B
\citep{yang2024qwen25technicalreport,chen2024huatuogpto1}. The analysis requires
tensor alignment, not merely architectural similarity: corresponding decoder
tensors must have matching shapes and parameter roles. This holds within
each released pair. For Gemma, we define the observed checkpoint movement as
\[
\Delta W_{\mathrm{all}} = W_{\mathrm{MedGemma}} - W_{\mathrm{Gemma}},
\]
and define the Qwen movement analogously. These are released checkpoint
differences, not learned editors, adapters, or optimized merge directions. Both
pairs are public, same-lineage, and tensor-aligned, so their updates can be
audited without inferring a training trajectory or matching unrelated model
families.

Because our benchmark inputs are text-only, the main interventions use the
shared decoder-side portion of this movement. Let \(\Delta W_{\mathrm{dec}}\)
denote the aligned decoder matrices from \(\Delta W_{\mathrm{all}}\). The
reference decoder path is
\[
W_{\mathrm{dec}}(t)=W_{\mathrm{Gemma}} + t\Delta W_{\mathrm{dec}}.
\]
Here \(t=0\) is the base decoder and \(t=1\) is the decoder-side reconstruction
of the specialist for text-only evaluation. MedGemma also contains multimodal
components; we analyze encoder-side changes descriptively, but the primary
behavioral audit targets the text-only decoder path.
For Qwen, the same construction uses Qwen2.5 at \(t=0\) and HuatuoGPT-o1 at
\(t=1\). We use the linear additive path because it exposes fractions of the
observed update and makes component masking and matched controls directly
comparable. It is a diagnostic coordinate, not a claim about the training
trajectory; curved paths such as SLERP remain useful robustness checks.

The Gemma decoder audit covers 238 aligned two-dimensional matrices containing
3.209B parameters (96.8\% of the 3.316B aligned matrix parameters); 81
encoder-side matrices containing 0.107B parameters are described separately.
The Qwen audit covers 196 aligned decoder projection matrices, approximately
6.53B non-embedding parameters. Embeddings, normalization parameters, and output
heads are outside these matrix-family interventions.

\subsection{Path Families}

We evaluate three path families. The first is the full decoder path above, which
serves as the reference reconstruction. It asks whether the observed decoder-side
update carries the measured benchmark movement.

The second is a scoped-from-base component path. For a selected decoder family or
bucket \(S\), we apply only that subset of the update:
\[
W_S(t)=W_{\mathrm{Gemma}} + t\Delta W_S,
\]
where \(\Delta W_S\subseteq\Delta W_{\mathrm{dec}}\). These paths test partial
sufficiency: whether a subset alone can reproduce the reference movement.

The third is an endpoint-anchored rollback path. Starting from the full decoder
update, we roll back only a selected family \(S\):
\[
W_{\mathrm{anch},S}(\alpha)
= W_{\mathrm{dec}}(1) - (1-\alpha)\Delta W_S.
\]
We run this sweep for attention and MLP with
\(\alpha\in\{0,0.25,0.5,0.75,1\}\). Thus \(\alpha=1\) is the full decoder update,
while smaller values move only the selected family back toward Gemma. Scoped
paths ask whether a family is sufficient by itself; anchored rollback asks what
changes when that family is removed from an otherwise complete specialist
update. Neither path is a minimal-circuit claim.

\subsection{Evaluation Views}

For both checkpoint pairs, the target-domain readout is the same 1,810-example public medical multiple-choice
composite from MedQA and medical MMLU subsets. We restrict the composite to rows
with auditable public gold labels compatible with a single scoring pipeline. The
paper therefore audits measured benchmark movement, not clinical reliability.

We pair this with the same public non-clinical suite of 7,325 variable-choice
multiple-choice examples from GPQA Diamond, MMLU-Pro, CommonsenseQA, TruthfulQA
MC1, and text-only Kaleidoscope. This suite is used first as an aggregate
endpoint check. MedGemma is slightly higher than Gemma overall on the aggregate
suite (+0.0086 accuracy), so we do not claim broad non-clinical forgetting.

For path diagnostics, we also define selected non-clinical regression and gain
slices. Regression slices are sources with at least 80 examples and a
MedGemma--Gemma endpoint drop of at least 0.02 accuracy, yielding 2,469 examples.
Gain slices are defined symmetrically, yielding 2,606 examples. These slices are
endpoint-conditioned diagnostics of this checkpoint pair, not population-level
estimates of all non-clinical behavior.

\begin{table}[t]
    \centering
    \footnotesize
    \setlength{\tabcolsep}{3pt}
    \begin{tabular}{@{}p{0.30\columnwidth}p{0.48\columnwidth}r@{}}
        \toprule
        View & Sources & Count \\
        \midrule
        Medical composite & MedQA; MMLU Clinical Knowledge; MMLU Professional Medicine & 1,810 \\
        Full non-clinical & GPQA Diamond; MMLU-Pro; CommonsenseQA; TruthfulQA MC1; Kaleidoscope text-only & 7,325 \\
        Selected regressions & Endpoint drops with at least 80 examples and gap at least 0.02 & 2,469 \\
        Selected gains & Endpoint gains with at least 80 examples and gap at least 0.02 & 2,606 \\
        \bottomrule
    \end{tabular}
    \caption{Evaluation views. The medical composite is the target-domain
    readout; the full non-clinical suite checks aggregate endpoint behavior;
    selected regression and gain slices are conditional diagnostics used for path
    analysis.}
    \label{tab:data-summary}
\end{table}

\begin{table}[t]
    \centering
    \footnotesize
    \setlength{\tabcolsep}{4pt}
    \begin{tabular}{@{}lcc@{}}
        \toprule
        Endpoint comparison & $\Delta$ & 95\% CI \\
        \midrule
        Medical composite & +0.085 & [ +0.062, +0.108 ] \\
        Selected-regression slices & -0.035 & [ -0.058, -0.015 ] \\
        Selected-gain slices & +0.063 & [ +0.032, +0.084 ] \\
        \bottomrule
    \end{tabular}
    \caption{Endpoint movements used to normalize the path readouts. Deltas are
    MedGemma minus Gemma.}
    \label{tab:endpoint-deltas}
\end{table}

\subsection{Normalized Readouts}

Each example is scored by presenting the question and candidate options and
choosing the option label with the highest model score under a fixed prompt
format. Medical examples have four options; the non-clinical suite supports
variable option counts. The primary metric is accuracy.

For any metric \(m\), let \(m_G=m(W_{\mathrm{Gemma}})\) and
\(m_M=m(W_{\mathrm{MedGemma}})\). Medical retention is
\[
\mathrm{Ret}(W)=\frac{m(W)-m_G}{m_M-m_G}.
\]
A value of 0 means no gain over Gemma, 1 means the full MedGemma endpoint gain,
and values above 1 can occur when a path point slightly exceeds the specialist
endpoint on the finite benchmark.

For a selected regression slice \(s\), recovery is
\[
\mathrm{Rec}_s(W)=\frac{m_s(W)-m_{s,M}}{m_{s,G}-m_{s,M}},
\]
where \(m_{s,G}>m_{s,M}\) by construction. A value of 1 matches Gemma on the
selected regressions, 0 matches MedGemma, and negative values are worse than the
specialist endpoint. We report the macro-average over selected regression
slices.

For selected gain slices, the analogous quantity measures how much of
MedGemma's endpoint gain has been acquired:
\[
\mathrm{Gain}_s(W)=\frac{m_s(W)-m_{s,G}}{m_{s,M}-m_{s,G}}.
\]
We macro-average this quantity across selected gain slices and keep it separate
from regression recovery because the endpoint directions are opposite.

Confidence intervals use 2,000 paired bootstrap resamples with seed 1729.
Medical metrics resample examples. Slice diagnostics resample examples within
the endpoint-selected diagnostic set; slice definitions remain fixed after
endpoint selection.
The prompt, option-scoring rule, bootstrap procedure, path grid, and control
construction are held fixed across pairs. Regression and gain slices are defined
separately within each pair because they are endpoint-conditioned by design.

\section{What Changed in the Released Update?}
\label{sec:delta-screen}

Before evaluating behavior, we first inspect the released checkpoint difference
as a tensor object. For each aligned matrix, we compute
\(\Delta W_i=W_i^M-W_i^G\) and summarize its Frobenius norm, relative change
\(\|\Delta W_i\|_F/\|W_i^G\|_F\), root-mean-square change, and singular-value
spectrum. These quantities are descriptive: they identify where the released
update is large or structured, but they do not establish that a component causes
benchmark behavior.

Across 884 shared state-dict keys, we find 319 aligned weight matrices: 238 in
the shared decoder and 81 in encoder-side multimodal components, with no
alignment failures. Because all benchmark inputs in this paper are text-only,
the behavioral interventions target the shared decoder. Encoder-side changes are
reported descriptively in the appendix.

\begin{figure}[t]
    \centering
    \begin{subfigure}{\linewidth}
        \centering
        \includegraphics[width=\linewidth]{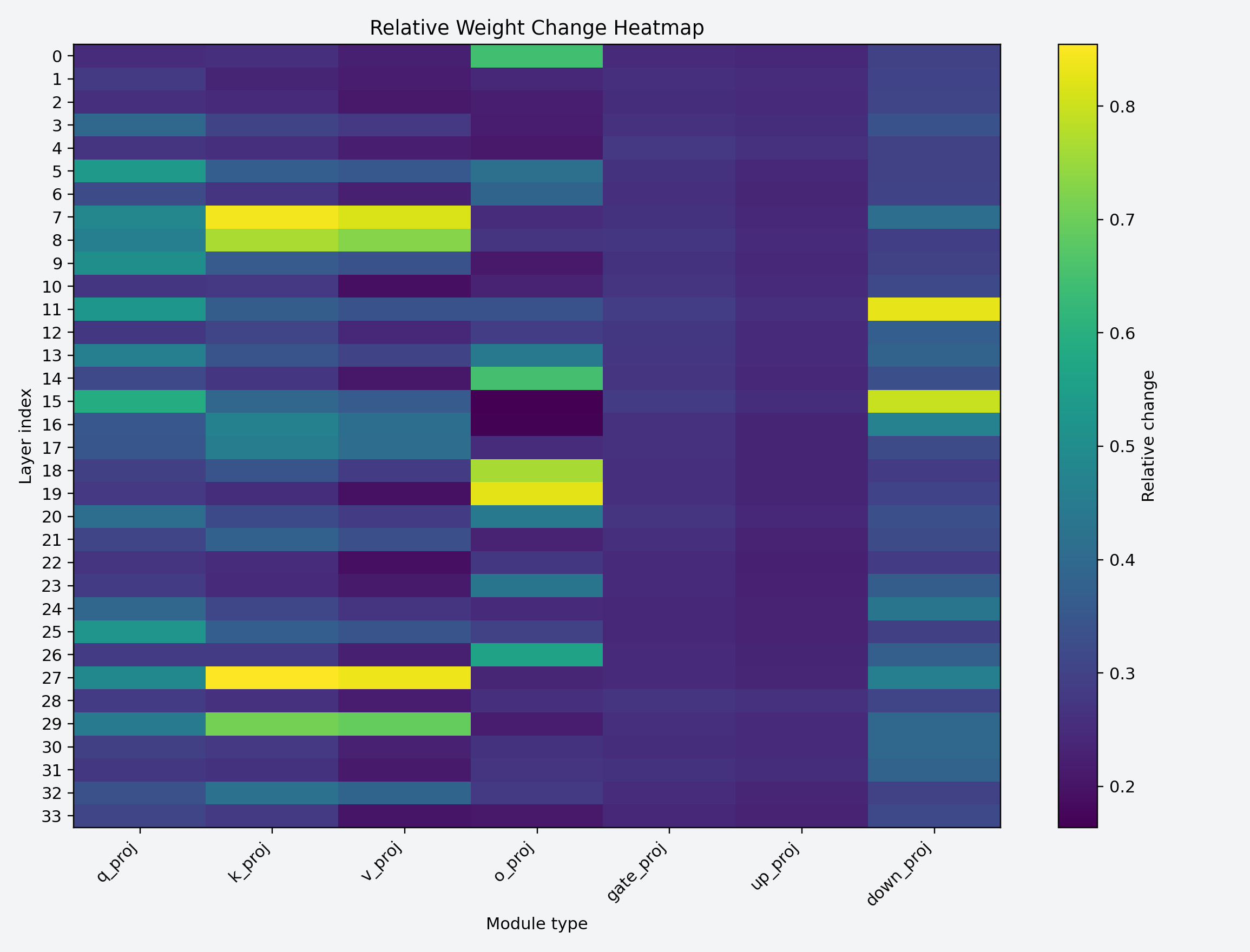}
        \caption{Layer--module relative change}
    \end{subfigure}

    \vspace{0.25em}
    \begin{subfigure}{\linewidth}
        \centering
        \includegraphics[width=\linewidth]{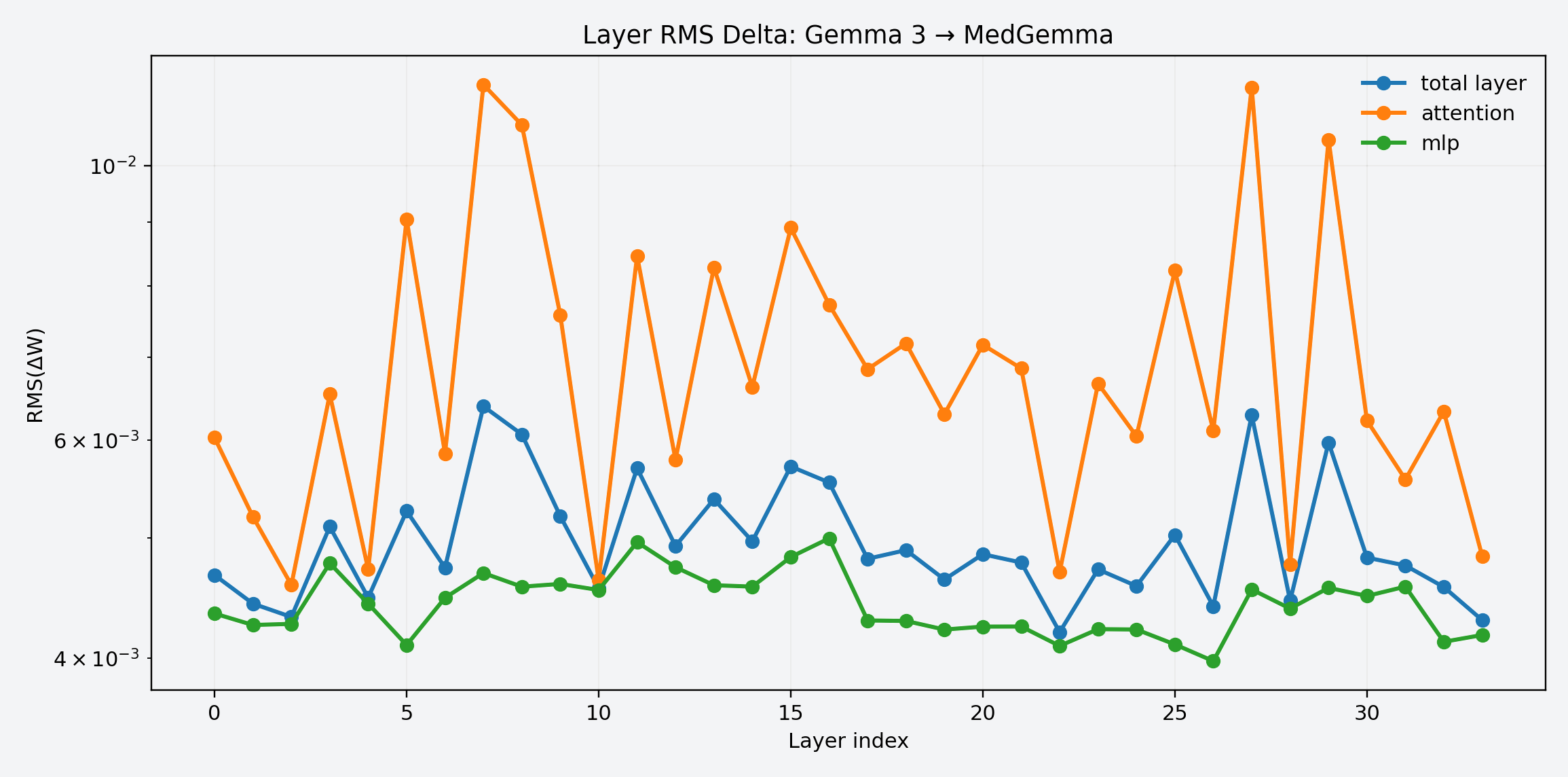}
        \caption{Layerwise RMS change}
    \end{subfigure}
    \caption{Decoder-side \(\Delta W\) screen. The released update is structured
    across layers and projection families rather than concentrated in a single
    obvious component. This screen motivates the tested buckets; behavioral
    claims come from path sweeps and controls.}
    \label{fig:delta-w-scoped}
\end{figure}

The screen reveals a broad, structured update rather than a tiny localized edit.
The appendix tables make this pattern concrete: decoder attention projections
show the clearest relative-change hotspots, while MLP projections account for a
large share of decoder-side update energy. Within the MLP family,
\texttt{down\_proj} is the most prominent projection by relative change, which
motivates the mixed attention-plus-\texttt{down\_proj} diagnostic. Spectral
summaries give the same caution from a different angle: update energy is not
concentrated in a handful of singular directions, and effective ranks remain
high across many decoder matrices (Appendix~\ref{sec:appendix-dw}). These facts
make simple one-family explanations suspect before any behavioral evaluation is
run.

The replication pair shows the same broad geometry at a larger scale. Its 196
aligned decoder projections comprise 112 attention and 84 MLP matrices. MLP
accounts for 87.4\% of projection parameters and 87.8\% of update energy, while
attention accounts for 12.6\% and 12.2\%, respectively. Mean effective ranks are
477.7 for attention and 1522.3 for MLP; within MLP, gate, up, and down projections
all remain high-rank (1471, 1354, and 1742). Thus the MLP-heavy screen is a prior
for intervention, not evidence that MLP uniquely carries the behavior.

This screen determines the probes used in the rest of the paper. The full
decoder path keeps all aligned decoder matrices and serves as the reference
reconstruction. The component paths test attention, MLP, a mixed
attention-plus-\texttt{down\_proj} bucket, and leave-one-family-out variants.
Matched random controls then ask whether bucket behavior reflects component
identity or simply the amount of update included. Anchored rollbacks ask the
complementary endpoint question: what happens if one family is removed from an
otherwise complete specialist update?

Thus the \(\Delta W\) screen is a routing step. It shows that the released
specialization update is broad enough to make simple localization suspect, but
it does not by itself explain behavior. The next section tests whether the
observed decoder update actually reconstructs benchmark movement.

\section{Does the Decoder Path Reconstruct Benchmark Movement?}
\label{sec:full-path}

We next ask whether the observed decoder-side update is behaviorally meaningful
for the text-only benchmarks. The full decoder path is
\[
W_{\mathrm{dec}}(t)=W_{\mathrm{Gemma}}+t\Delta W_{\mathrm{dec}}.
\]
This experiment is the reference reconstruction for the audit. The endpoint
\(t=1\) checks whether the aligned decoder update accounts for the measured
text-only endpoint movement; the trajectory between endpoints shows how medical
and selected non-clinical behavior enter as the released update is introduced.

\begin{figure*}[t]
    \centering
    \includegraphics[width=0.98\textwidth]{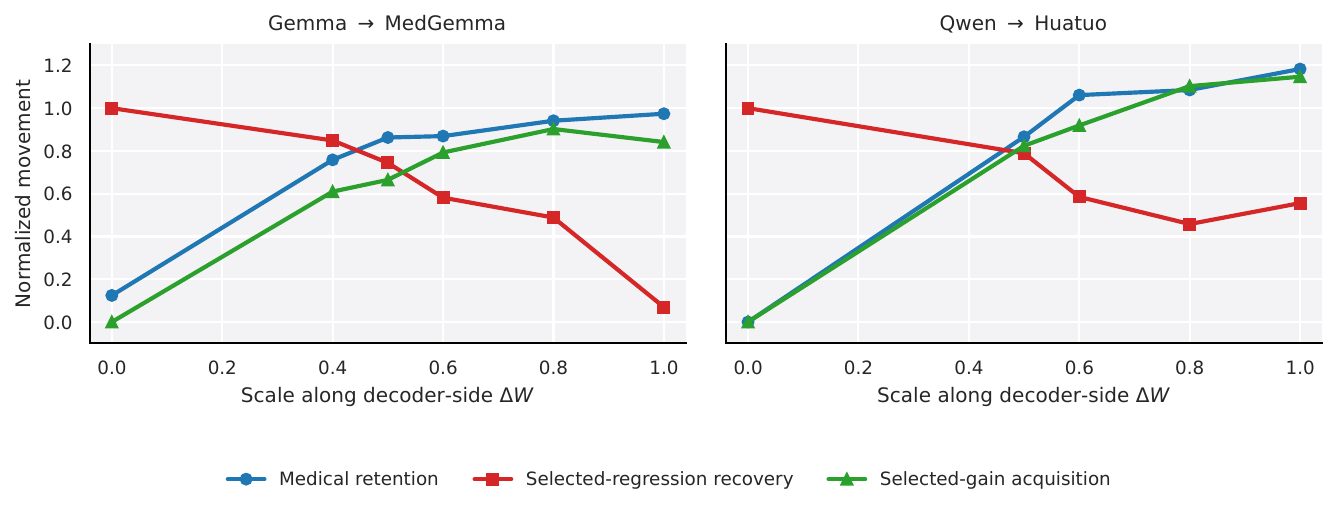}
    \caption{Full decoder-side paths for both checkpoint pairs. Each observed
    update reconstructs the medical endpoint movement while carrying mixed
    selected off-domain movements. The Qwen curves connect only evaluated path
    points. Selected slices are endpoint-conditioned diagnostics, not estimates
    of broad non-clinical forgetting.}
    \label{fig:unified-retention-recovery}
\end{figure*}

The full decoder path strongly reconstructs the measured medical benchmark
movement. At the specialist endpoint, it reaches 0.974 normalized medical
retention relative to the standalone Gemma\(\rightarrow\)MedGemma medical gain
(Table~\ref{tab:endpoint-deltas}). This establishes the aligned decoder update
as the relevant substrate for the text-only audit. Component probes are meaningful only because this audited decoder delta first
recovers the endpoint behavior. We do not interpret this as a clinical result or
as a mechanism; it is the coordinate system against which the component and
control experiments are compared.

The trajectory reveals structure that the endpoint score hides. Medical
retention rises early and remains high through the later part of the path; for
example, the 0.50 point retains 0.863 of the medical gain while preserving 0.744
selected-regression recovery, and the 0.60 point retains 0.869 medical retention
with 0.793 selected-gain acquisition. The bootstrap bands in
Appendix Figure~\ref{fig:full-path-ci-main} support the qualitative medical trend, while
also showing that selected regression and gain diagnostics are wider and should
be read directionally. This does not imply broad forgetting: on the full
non-clinical suite, MedGemma is slightly higher than Gemma overall (+0.0086
accuracy; 95\% CI [-0.002, +0.020]). The selected-regression view is therefore a
post-hoc diagnostic of slices that moved downward at the endpoint, not a global
estimate of non-clinical capability.

The replication yields the same central result despite a different trajectory.
The Qwen decoder path reaches 1.183 normalized medical retention at
\(t=1\) (95\% CI [0.962, 1.547]); values above 1 mean that this decoder
reconstruction slightly exceeds the finite-sample specialist endpoint, not
``118.3\% capability.'' Medical retention rises from 0.866 at \(t=0.5\) to 1.085
at \(t=0.8\), while selected-regression recovery falls from 0.789 to 0.458 and
selected-gain acquisition rises from 0.824 to 1.103. The Qwen endpoint movements
are +0.0453 medical accuracy (95\% CI [0.0254, 0.0641]), -0.0617 on its selected
regressions, and +0.0554 on its selected gains.

The selected-gain view completes the picture. Some non-clinical slices improve
rather than regress, and their gain acquisition also rises along the decoder
path. Across both pairs, the specialist update is not well described as a one-dimensional
tradeoff between medical ability and general ability. It carries the target
medical movement together with heterogeneous off-domain changes.

We next ask whether a smaller component family explains the reconstructed
movement, using scoped paths, matched controls, and anchored rollbacks.

\section{Can Coarse Components Explain the Movement?}
\label{sec:components}

MLP is the strongest coarse bucket, but that alone does not establish
localization. We therefore ask whether the reconstructed movement can be
attributed to a smaller component family using three probes: scoped-from-base
component paths, matched random controls, and
endpoint-anchored rollbacks. These are coarse component-family diagnostics, not
circuit-localization claims.

\begin{figure}[t]
    \centering
    \includegraphics[width=\linewidth]{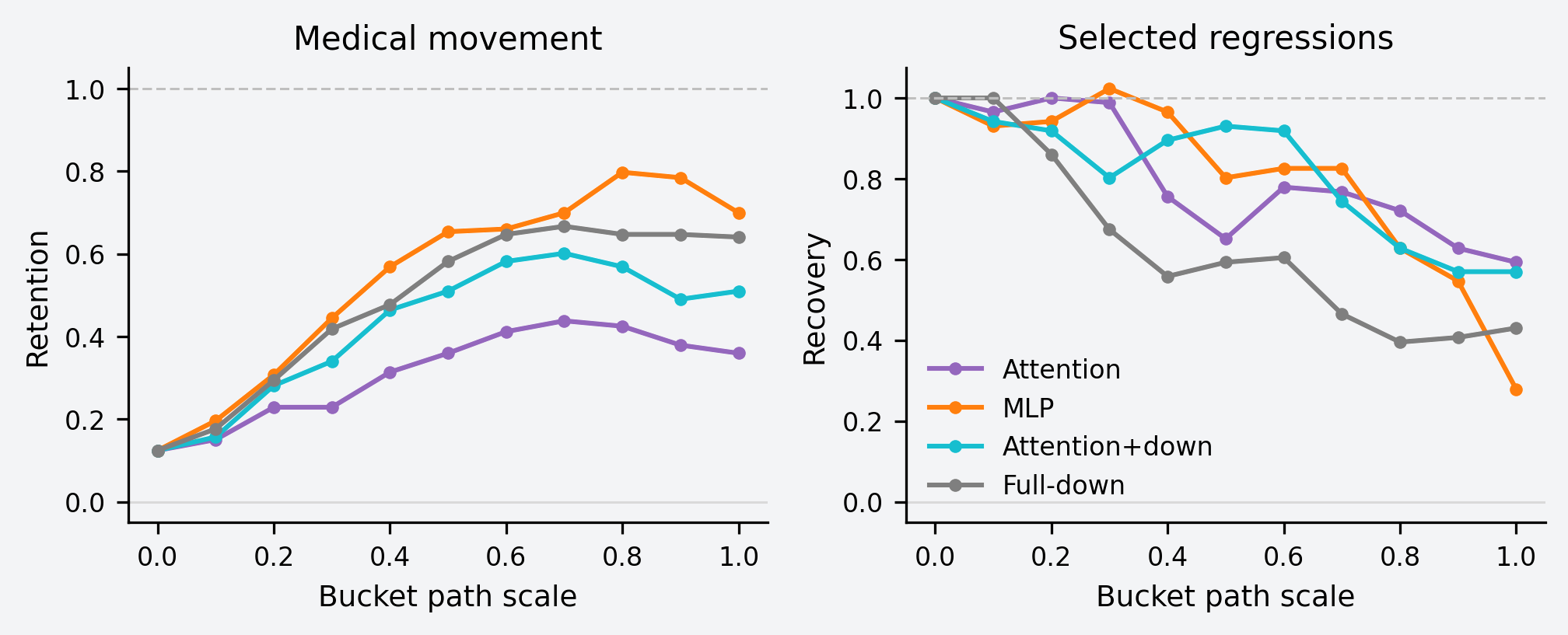}
    \caption{Scoped component-family paths show partial sufficiency, not
    localization. The MLP family is strongest among scoped buckets, but no
    tested bucket matches the full decoder path and the strongest bucket is also
    large.}
    \label{fig:bucket-curves}
\end{figure}

Scoped paths show partial sufficiency but not a complete explanation. The MLP
path is strongest among the tested buckets, reaching 0.797 medical retention at
its best point. Attention alone is weaker, and the
attention-plus-\texttt{down\_proj} bucket reaches 0.601 retention. The
full-minus-\texttt{down\_proj} path retains more medical movement than the mixed
attention-plus-\texttt{down\_proj} bucket, so \texttt{down\_proj} is not
necessary in isolation. The scoped-path result is therefore suggestive but
limited: broad families can reproduce parts of the medical movement, but no
tested bucket matches the full decoder path.

\begin{table}[t]
    \centering
    \footnotesize
    \setlength{\tabcolsep}{5pt}
    \begin{tabular}{lrrr}
\toprule
Bucket & Best scale & Med. ret. & Reg. rec. \\
\midrule
Attention & 0.70 & 0.438 & 0.767 \\
MLP & 0.80 & 0.797 & 0.628 \\
Attention+down & 0.70 & 0.601 & 0.744 \\
Full-down & 0.70 & 0.667 & 0.465 \\
\bottomrule
\end{tabular}

    \caption{Best observed points for scoped decoder-side component-family
    paths. Medical retention is normalized by the endpoint medical gain;
    selected-regression recovery is endpoint-conditioned.}
    \label{tab:bucket-best}
\end{table}

The strongest scoped result also raises an obvious concern: the MLP bucket is
large. We therefore compare structured buckets with matched random controls that
match matrix count, parameter count, or \(\Delta W\)-energy. For each target and
matching rule, we run 10 seeded draws. These controls are still coarse nulls,
but they test whether a structured bucket is clearly distinguished from similarly
sized or similarly energetic subsets of the same released update.

\begin{table}[t]
    \centering
    \footnotesize
    \setlength{\tabcolsep}{4pt}
    \resizebox{\linewidth}{!}{\begin{tabular}{llrr}
\toprule
Diagnostic & Point & Med. ret. & Reg. rec. \\
\midrule
MLP scoped best & $t=0.80$ & 0.797 & 0.628 \\
MLP param. ctrl (10) & $t=1.00$ & 0.814$\pm$0.067 & 0.236$\pm$0.175 \\
MLP $\Delta$E ctrl (10) & $t=1.00$ & 0.563$\pm$0.100 & 0.262$\pm$0.242 \\
Attn+down scoped best & $t=0.70$ & 0.601 & 0.744 \\
Attn+down matrix ctrl (10) & $t=1.00$ & 0.703$\pm$0.112 & 0.388$\pm$0.159 \\
Attn+down param. ctrl (10) & $t=1.00$ & 0.417$\pm$0.086 & 0.585$\pm$0.166 \\
Attention rollback & $\alpha=0$ & 0.699 & 0.279 \\
MLP rollback & $\alpha=0$ & 0.359 & 0.593 \\
\bottomrule
\end{tabular}
}
    \caption{Controls and anchored rollbacks test whether component sufficiency
    becomes explanation. The table distinguishes scoped-best path points from
    controls at \(t=1\) and full rollbacks at \(\alpha=0\). Control entries are
    mean $\pm$ standard deviation over 10 seeds.}
    \label{tab:component-checks}
\end{table}

Matched controls weaken a unique-family interpretation: if random matched
subsets approach a structured bucket, the bucket is not a unique semantic carrier
at this granularity. At \(t=1\), parameter-count-matched MLP
controls average 0.814 medical retention, exceeding the structured MLP value
(0.699), although their selected-regression recovery remains close. For
attention-plus-\texttt{down\_proj}, matrix-count controls average higher medical
retention than the structured bucket, while parameter-count controls approach its
selected-regression recovery. These results do not imply that component identity
is irrelevant, but they show that bucket performance depends substantially on
how much of the observed update is included and on which axis is matched
(Appendix Figure~\ref{fig:appendix-bucket-control-scatter}). Component-family
paths are therefore partial sufficiency probes, not unique explanations.

The replication makes this control result harder to dismiss as pair-specific.
For Qwen--HuatuoGPT, MLP reaches 1.024 medical retention at \(t=1\), but the
MLP parameter- and energy-matched controls reach 1.115 and 1.098 on average.
Attention reaches only 0.341, while attention-plus-\texttt{down\_proj} reaches
0.841. These are comparisons at the same endpoint scale; we do not compare a
bucket's best intermediate point against a control evaluated only at \(t=1\).

\begin{table}[t]
    \centering
    \scriptsize
    \setlength{\tabcolsep}{2.5pt}
    \resizebox{\linewidth}{!}{\begin{tabular}{@{}lrr@{}}
\toprule
Diagnostic & Gemma--MedGemma & Qwen--Huatuo \\
\midrule
Medical endpoint gain & +0.0845 & +0.0453 \\
Full decoder retention & 0.974 & 1.183 \\
MLP retention at $t=1$ & 0.699 & 1.024 \\
MLP parameter control & 0.814 & 1.115 \\
MLP $\Delta W$-energy control & 0.563 & 1.098 \\
Attention rollback: med./reg. & 0.699 / 0.279 & 0.756 / 0.387 \\
MLP rollback: med./reg. & 0.359 / 0.593 & 0.293 / 0.986 \\
\bottomrule
\end{tabular}
}
    \caption{Cross-pair audit summary. Control entries are mean medical retention
    at \(t=1\); rollback entries report medical retention / selected-regression
    recovery after fully removing the named family.}
    \label{tab:cross-pair-summary}
\end{table}

Endpoint-anchored rollbacks provide the complementary test. Instead of asking
whether a family can reproduce movement from the base model, they ask what
happens when one family is rolled back from an otherwise complete specialist
update.

\begin{figure}[t]
    \centering
    \includegraphics[width=\linewidth]{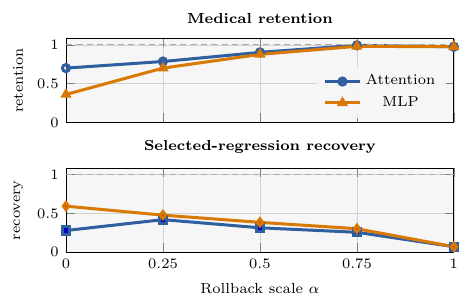}
    \caption{Endpoint-anchored rollback sweeps show that no family is a clean
    repair knob. MLP rollback recovers selected regressions more than attention
    rollback, but at a much larger medical-retention cost.}
    \label{fig:anchor-sweep-main}
\end{figure}

The rollback result again argues against a simple component knob. Full MLP
rollback recovers more of the selected regressions than full attention rollback
(0.593 versus 0.279), but it also costs far more medical retention (0.359 versus
0.699). Partial rollbacks do not produce a clean frontier where regressions are
recovered while medical retention is preserved. Thus the endpoint behavior is
not organized as a monotone sum of independently helpful component families.
Qwen shows the same tradeoff more sharply: full MLP rollback retains 0.293 of
the medical gain while recovering 0.986 of selected regressions, whereas
attention rollback retains 0.756 medical and recovers 0.387. Neither pair offers
a clean family-level repair that preserves the target gain while reversing the
selected regressions.

Taken together, these experiments make the component result mostly negative but
useful. The medical benchmark movement is reproducible along the full decoder
path, and broad families can reproduce parts of it. However, matched controls
and anchored rollbacks prevent a stronger localization claim: at the tested
granularity, no component family uniquely explains the movement.

\paragraph{Empirical summary.}
The main results are therefore threefold. First, the medical benchmark movement
is carried by the observed decoder-side updates: both full paths reconstruct the
measured medical gain. Second, each update carries heterogeneous
off-domain movement, with selected regressions, selected gains, and a nearly
flat aggregate non-clinical endpoint. Third, coarse component families are useful
probes but weak explanations: MLP is the strongest scoped bucket, yet matched
controls and anchored rollbacks prevent a unique family-level localization
claim, and this negative conclusion replicates across both checkpoint pairs.

\section{Discussion}

Same-lineage specialist releases deserve an analysis layer between endpoint
evaluation and circuit discovery. Endpoint evaluation treats each specialist as
a model; a weight-delta audit treats the aligned generalist-to-specialist
difference as an object with internal structure. Across two medical checkpoint
pairs, the full decoder update reconstructs the measured target movement, while
selected off-domain gains and regressions travel differently along the path.
This is stronger than an endpoint comparison but weaker than a mechanistic
explanation: the paths establish an auditable substrate, not the features,
circuits, training examples, or historical trajectory that caused it.

The cross-pair result also sharpens how component evidence should be read. MLP is
the strongest broad component family in both pairs, consistent with the descriptive
energy screen and prior work on feed-forward knowledge storage. Yet MLP is also
the dominant parameter and update-energy family, and matched random subsets can
equal or exceed it. The reproducible conclusion is therefore not ``medical
specialization is in the MLPs.'' It is that MLP provides a useful intervention
prior whose apparent sufficiency does not survive controls as unique semantic
localization. Parameter mass and update energy can masquerade as explanation.

Rollback makes the practical consequence visible. Removing MLP recovers more of
the selected regressions in both pairs, but it also removes much more medical
movement than attention rollback. No tested family is a clean repair knob. In
practice, the audit is better suited to release comparison, regression triage,
and prioritizing finer-grained follow-up than to editing a deployed model. A
promising bucket should earn stronger evidence through controls, multiple
checkpoint pairs, finer causal interventions, and open-ended behavioral tests.

Finally, the non-clinical views remain conditional diagnostics. Their source
composition differs semantically, the gain/regression slices are selected from
each pair's endpoints, and multiple-choice accuracy does not capture calibration
or generation quality. A useful next step is to pair the weight-space audit with
open-ended clinical tasks and blinded judge ensembles, explicitly measuring
judge variance rather than replacing auditable labels with a single evaluator.
The present result is deliberately narrower: for two released, tensor-aligned
medical specializations, full-path reconstruction replicates, whereas a unique
coarse-family explanation does not.

\section{Conclusion}

We audit two public generalist-to-medical-specialist weight deltas. In both
Gemma--MedGemma and Qwen--HuatuoGPT, the aligned decoder path reconstructs the
measured medical benchmark gain while carrying heterogeneous selected
non-clinical movement. MLP is the strongest broad component family, but matched
controls and endpoint-anchored rollbacks prevent unique localization: size and
update energy explain substantial apparent sufficiency, and removing MLP trades
regression recovery for medical retention. Endpoint gains should therefore be
paired with update-level audits before component or repair narratives are
inferred; these results are neither clinical validation nor circuit explanation.

\section{Limitations}

This study covers two public, tensor-aligned medical checkpoint pairs, not a
general theory of post-training. Both audited intervention paths operate on
decoder-only language-model backbones specialized for medicine; other
architectures, domains, sizes, and training
recipes may organize their updates differently, and the protocol requires
comparable endpoint tensors. The evaluation is text-only and multiple-choice. It
does not measure free-form clinical generation, calibration, abstention,
multimodal use, robustness, harmful-answer behavior, or clinical utility. Public
benchmark overlap may exist, so gains are measured benchmark movement rather
than out-of-distribution medical generalization.

The additive weight path is a counterfactual audit coordinate, not the historical
training trajectory or a deployment recommendation. Alternative geometries such
as SLERP could yield different intermediate behavior. Embeddings, normalization
parameters, output heads, and Gemma's vision-side parameters are outside the
primary matrix-family interventions. Component buckets test coarse partial
sufficiency rather than circuits, features, token-level mechanisms, or minimal
subsets. Ten-seed matched controls for both pairs are informative
but not exhaustive null distributions. Finally, regression and gain slices are
selected separately from each pair's endpoints and combine semantically varied
benchmarks; they are diagnostic views, not population estimates of general
capability or forgetting.

\section*{Acknowledgments}

We thank Sri Gadde, Krishnaram Kenthapadi, Raefer Gabriel,
Anurag Dwarkanath, and Kiran Rama from the Oracle Health leadership team
for their guidance, thoughtful feedback, and continued support throughout
this work. We also gratefully acknowledge Oracle Health AI for supporting
this research and providing the computational resources that enabled the
experiments and analyses presented in this paper.

\section*{Ethics Statement}

This work analyzes public checkpoints and public benchmark examples, and uses no patient records, protected health information, or human-subject data. The medical setting is sensitive: exam-style multiple-choice gains should not be read as clinical reliability, safe advice generation, calibrated uncertainty, or readiness for patient- or clinician-facing use. The interpolation, bucket, rollback, and neuron-diagnostic analyses are retrospective audit probes, not certification, repair, or deployment methods. Weight-space analysis can be dual-use because it may inform capability modification; we mitigate this by keeping claims benchmark-scoped, reporting negative controls and limitations with the positive results, not releasing modified checkpoints, and framing the contribution as audit evidence rather than clinical guidance.

\bibliography{references}

\appendix

\section{Additional Results and Diagnostics}
\FloatBarrier
\makeatletter
\setlength{\@dblfptop}{0pt}
\setlength{\@dblfpsep}{8pt plus 2pt minus 2pt}
\setlength{\@dblfpbot}{0pt plus 1fil}
\makeatother

\subsection{Evaluation Details and Endpoint Movements}

The primary medical composite contains only rows with auditable public labels
compatible with the scoring protocol. The non-clinical suite and the selected
regression/gain diagnostics follow the definitions in Section~\ref{sec:setup-data}.
Table~\ref{tab:appendix-eval-views-v5} summarizes the evaluation views, and
Table~\ref{tab:appendix-endpoint-deltas-v5} reports the endpoint
movements that define the normalized readouts.

\begin{table}[H]
    \centering
    \scriptsize
    \setlength{\tabcolsep}{3pt}
    
\begin{tabular}{@{}p{0.30\linewidth}p{0.54\linewidth}r@{}}
\toprule
View & Sources / rule & Count \\
\midrule
Medical composite & MedQA; MMLU Clinical Knowledge; MMLU Professional Medicine & 1,810 \\
Full non-clinical & GPQA Diamond; MMLU-Pro; CommonsenseQA; TruthfulQA MC1; Kaleidoscope text-only & 7,325 \\
Selected regressions & Endpoint drops with at least 80 examples and gap at least 0.02 & 2,469 \\
Selected gains & Endpoint gains with at least 80 examples and gap at least 0.02 & 2,606 \\
\bottomrule
\end{tabular}

    \caption{Evaluation views used in the audit. Selected regression and
    gain views are endpoint-conditioned diagnostics, not population estimates of
    all non-clinical behavior.}
    \label{tab:appendix-eval-views-v5}
\end{table}

\begin{table}[H]
    \centering
    \footnotesize
    \setlength{\tabcolsep}{5pt}
    \begin{tabular}{@{}lrr@{}}
        \toprule
        View & $\Delta$ & 95\% CI \\
        \midrule
        Medical & +0.085 & [+0.062, +0.108] \\
        Selected regressions & -0.035 & [-0.058, -0.015] \\
        Selected gains & +0.063 & [+0.032, +0.084] \\
        \bottomrule
    \end{tabular}
    \caption{Endpoint movements used to normalize the path readouts. Deltas are
    MedGemma minus Gemma.}
    \label{tab:appendix-endpoint-deltas-v5}
\end{table}

\begin{figure}[H]
    \centering
    \includegraphics[width=0.92\linewidth]{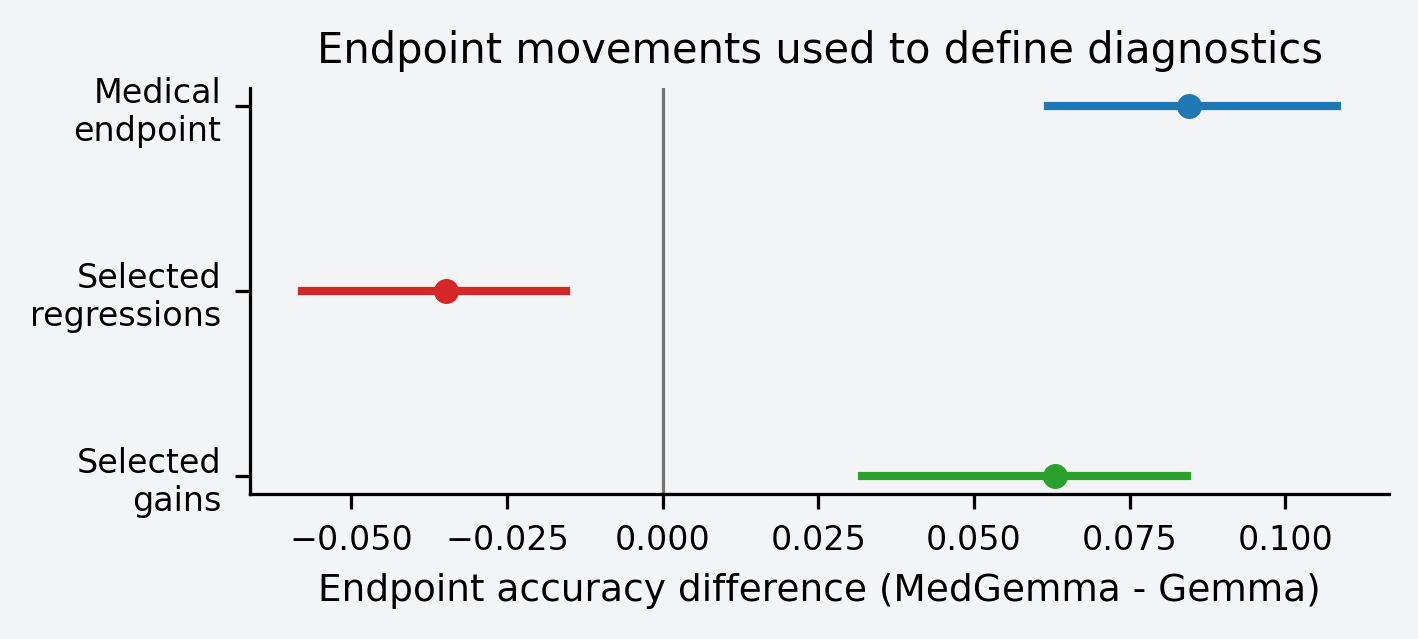}
    \caption{Endpoint movement summary with bootstrap intervals. The full
    non-clinical aggregate is used descriptively in the main text and is not the
    denominator for the selected-slice readouts.}
    \label{fig:appendix-endpoint-deltas-v5}

\end{figure}

\begin{table*}[!tbp]
    \centering
    \scriptsize
    \setlength{\tabcolsep}{4pt}
    \begin{tabular}{lrrrrr}
\toprule
Source & Count & Gemma & MedGemma & $\Delta$ & 95\% CI \\
\midrule
MedQA (USMLE 4-option) & 1,273 & 0.469 & 0.551 & +0.082 & [+0.053, +0.109] \\
MMLU Clinical Knowledge & 265 & 0.604 & 0.645 & +0.042 & [-0.019, +0.098] \\
MMLU Professional Medicine & 272 & 0.529 & 0.669 & +0.140 & [+0.081, +0.199] \\
\midrule
Composite & 1,810 & 0.498 & 0.582 & +0.085 & [+0.062, +0.108] \\
\bottomrule
\end{tabular}

    \caption{Medical endpoint performance by source. This table makes explicit
    that the composite gain is positive across the three medical sources,
    while the composite is example-weighted and therefore MedQA-dominated.}
    \label{tab:appendix-medical-sources-v5}
\end{table*}

\begin{table}[H]
    \centering
    \footnotesize
    \setlength{\tabcolsep}{4pt}
    \resizebox{\columnwidth}{!}{%
\begin{tabular}{lrrrrr}
\toprule
View & Count & Gemma & MedGemma & $\Delta$ & 95\% CI \\
\midrule
Full non-clinical & 7,325 & 0.410 & 0.418 & +0.009 & [-0.002, +0.020] \\
\bottomrule
\end{tabular}%
}

    \caption{Full non-clinical endpoint aggregate. The small positive aggregate
    movement motivates treating selected regressions as diagnostics rather than
    broad forgetting.}
    \label{tab:appendix-nonclinical-aggregate-v5}
\end{table}

\begin{table*}[!tbp]
    \centering
    \scriptsize
    \setlength{\tabcolsep}{2.5pt}
    \begin{subtable}{0.49\textwidth}
        \centering
        \begin{tabular}{@{}lrrrr@{}}
            \toprule
            Source & Count & Gemma & MedGemma & $\Delta$ \\
            \midrule
            commonsenseqa & 1,221 & 0.704 & 0.672 & -0.032 \\
            gpqa\_diamond & 198 & 0.278 & 0.232 & -0.045 \\
            kaleidoscope\_ar & 100 & 0.460 & 0.450 & -0.010 \\
            kaleidoscope\_bn & 100 & 0.380 & 0.410 & +0.030 \\
            kaleidoscope\_de & 100 & 0.520 & 0.560 & +0.040 \\
            kaleidoscope\_en & 100 & 0.500 & 0.470 & -0.030 \\
            kaleidoscope\_es & 100 & 0.520 & 0.570 & +0.050 \\
            kaleidoscope\_fa & 100 & 0.280 & 0.290 & +0.010 \\
            kaleidoscope\_fr & 100 & 0.230 & 0.290 & +0.060 \\
            kaleidoscope\_hi & 100 & 0.290 & 0.300 & +0.010 \\
            kaleidoscope\_hr & 100 & 0.270 & 0.250 & -0.020 \\
            kaleidoscope\_hu & 100 & 0.320 & 0.260 & -0.060 \\
            kaleidoscope\_lt & 100 & 0.550 & 0.570 & +0.020 \\
            kaleidoscope\_nl & 100 & 0.360 & 0.410 & +0.050 \\
            kaleidoscope\_pt & 100 & 0.690 & 0.730 & +0.040 \\
            kaleidoscope\_ru & 100 & 0.230 & 0.330 & +0.100 \\
            \bottomrule
        \end{tabular}
    \end{subtable}\hfill
    \begin{subtable}{0.49\textwidth}
        \centering
        \begin{tabular}{@{}lrrrr@{}}
            \toprule
            Source & Count & Gemma & MedGemma & $\Delta$ \\
            \midrule
            kaleidoscope\_sr & 100 & 0.250 & 0.230 & -0.020 \\
            kaleidoscope\_uk & 100 & 0.350 & 0.380 & +0.030 \\
            mmlu\_pro\_biology & 250 & 0.596 & 0.580 & -0.016 \\
            mmlu\_pro\_business & 250 & 0.252 & 0.248 & -0.004 \\
            mmlu\_pro\_chemistry & 250 & 0.184 & 0.172 & -0.012 \\
            mmlu\_pro\_computer\_science & 250 & 0.304 & 0.272 & -0.032 \\
            mmlu\_pro\_economics & 250 & 0.472 & 0.452 & -0.020 \\
            mmlu\_pro\_engineering & 250 & 0.240 & 0.228 & -0.012 \\
            mmlu\_pro\_health & 250 & 0.352 & 0.412 & +0.060 \\
            mmlu\_pro\_history & 250 & 0.364 & 0.372 & +0.008 \\
            mmlu\_pro\_law & 250 & 0.240 & 0.232 & -0.008 \\
            mmlu\_pro\_math & 250 & 0.172 & 0.144 & -0.028 \\
            mmlu\_pro\_other & 250 & 0.308 & 0.260 & -0.048 \\
            mmlu\_pro\_philosophy & 250 & 0.276 & 0.352 & +0.076 \\
            mmlu\_pro\_physics & 250 & 0.152 & 0.184 & +0.032 \\
            mmlu\_pro\_psychology & 250 & 0.452 & 0.484 & +0.032 \\
            truthfulqa\_mc1 & 806 & 0.465 & 0.557 & +0.092 \\
            \bottomrule
        \end{tabular}
    \end{subtable}
    \caption{Exact full non-clinical source composition and endpoint movements.}
    \label{tab:appendix-nonclinical-sources-v5}
\end{table*}

\begin{table*}[!tbp]
    \centering
    \scriptsize
    \begin{subtable}{0.48\textwidth}
        \centering
        \resizebox{\linewidth}{!}{\begin{tabular}{lrrrr}
\toprule
Source & Count & Gemma & MedGemma & $\Delta$ \\
\midrule
kaleidoscope\_hu & 100 & 0.320 & 0.260 & -0.060 \\
mmlu\_pro\_other & 250 & 0.308 & 0.260 & -0.048 \\
gpqa\_diamond & 198 & 0.278 & 0.232 & -0.045 \\
mmlu\_pro\_computer\_science & 250 & 0.304 & 0.272 & -0.032 \\
commonsenseqa & 1,221 & 0.704 & 0.672 & -0.032 \\
kaleidoscope\_en & 100 & 0.500 & 0.470 & -0.030 \\
mmlu\_pro\_math & 250 & 0.172 & 0.144 & -0.028 \\
kaleidoscope\_hr & 100 & 0.270 & 0.250 & -0.020 \\
\bottomrule
\end{tabular}
}
        \caption{Selected regressions}
    \end{subtable}
    \hfill
    \begin{subtable}{0.48\textwidth}
        \centering
        \resizebox{\linewidth}{!}{\begin{tabular}{lrrrr}
\toprule
Source & Count & Gemma & MedGemma & $\Delta$ \\
\midrule
kaleidoscope\_ru & 100 & 0.230 & 0.330 & +0.100 \\
truthfulqa\_mc1 & 806 & 0.465 & 0.557 & +0.092 \\
mmlu\_pro\_philosophy & 250 & 0.276 & 0.352 & +0.076 \\
mmlu\_pro\_health & 250 & 0.352 & 0.412 & +0.060 \\
kaleidoscope\_fr & 100 & 0.230 & 0.290 & +0.060 \\
kaleidoscope\_nl & 100 & 0.360 & 0.410 & +0.050 \\
kaleidoscope\_es & 100 & 0.520 & 0.570 & +0.050 \\
kaleidoscope\_de & 100 & 0.520 & 0.560 & +0.040 \\
kaleidoscope\_pt & 100 & 0.690 & 0.730 & +0.040 \\
mmlu\_pro\_physics & 250 & 0.152 & 0.184 & +0.032 \\
mmlu\_pro\_psychology & 250 & 0.452 & 0.484 & +0.032 \\
kaleidoscope\_uk & 100 & 0.350 & 0.380 & +0.030 \\
kaleidoscope\_bn & 100 & 0.380 & 0.410 & +0.030 \\
\bottomrule
\end{tabular}
}
        \caption{Selected gains}
    \end{subtable}
    \caption{Endpoint-selected diagnostic slices. Selection uses source-level
    endpoint gaps of at least 0.02 accuracy and at least 80 examples.}
    \label{tab:appendix-selected-slices-v5}
\end{table*}

All model evaluations use the same multiple-choice scoring rule: each example is
presented with the question stem and answer options, and the predicted answer is
the option label with the highest model score. Medical examples use four options;
the non-clinical suite supports variable option counts. Bootstrap intervals use
2,000 paired resamples with seed 1729; medical metrics resample examples, while
slice diagnostics resample examples within the fixed endpoint-selected sets.

\subsection{Full Decoder-Side Path Diagnostics}

Table~\ref{tab:appendix-full-path-grid-v5} gives the full decoder-side path grid
used for the main path analysis. Normalized retention is computed against the
standalone endpoint evaluations in Table~\ref{tab:appendix-endpoint-deltas-v5}.
A separately evaluated \(t=0\) reconstruction serves only as a consistency check
and is omitted from the grid; it is not used as the normalization denominator.
The selected-regression and selected-gain figures show that
endpoint-conditioned slices move heterogeneously, which is why the paper reports
regression recovery and gain acquisition as separate readouts.

\begin{table}[H]
    \centering
    \scriptsize
    \setlength{\tabcolsep}{4pt}
    \begin{tabular}{rrrrr}
\toprule
$t$ & Med. acc. & Med. ret. & Reg. rec. & Gain acq. \\
\midrule
0.10 & 0.5193 & 0.255 & 0.953 & 0.220 \\
0.20 & 0.5293 & 0.373 & 0.965 & 0.354 \\
0.25 & 0.5365 & 0.458 & 1.047 & 0.445 \\
0.30 & 0.5448 & 0.556 & 0.895 & 0.506 \\
0.40 & 0.5619 & 0.758 & 0.849 & 0.610 \\
0.50 & 0.5707 & 0.863 & 0.744 & 0.665 \\
0.60 & 0.5713 & 0.869 & 0.581 & 0.793 \\
0.70 & 0.5751 & 0.915 & 0.651 & 0.866 \\
0.80 & 0.5773 & 0.941 & 0.488 & 0.902 \\
0.90 & 0.5818 & 0.993 & 0.372 & 0.866 \\
1.00 & 0.5801 & 0.974 & 0.070 & 0.841 \\
\bottomrule
\end{tabular}

    \caption{Full decoder-side path grid. Gain acquisition is the acquired
    fraction of selected endpoint gains; selected-regression recovery is computed
    on endpoint-regression slices.}
    \label{tab:appendix-full-path-grid-v5}
\end{table}

\begin{figure}[H]
    \centering
    \includegraphics[width=\linewidth]{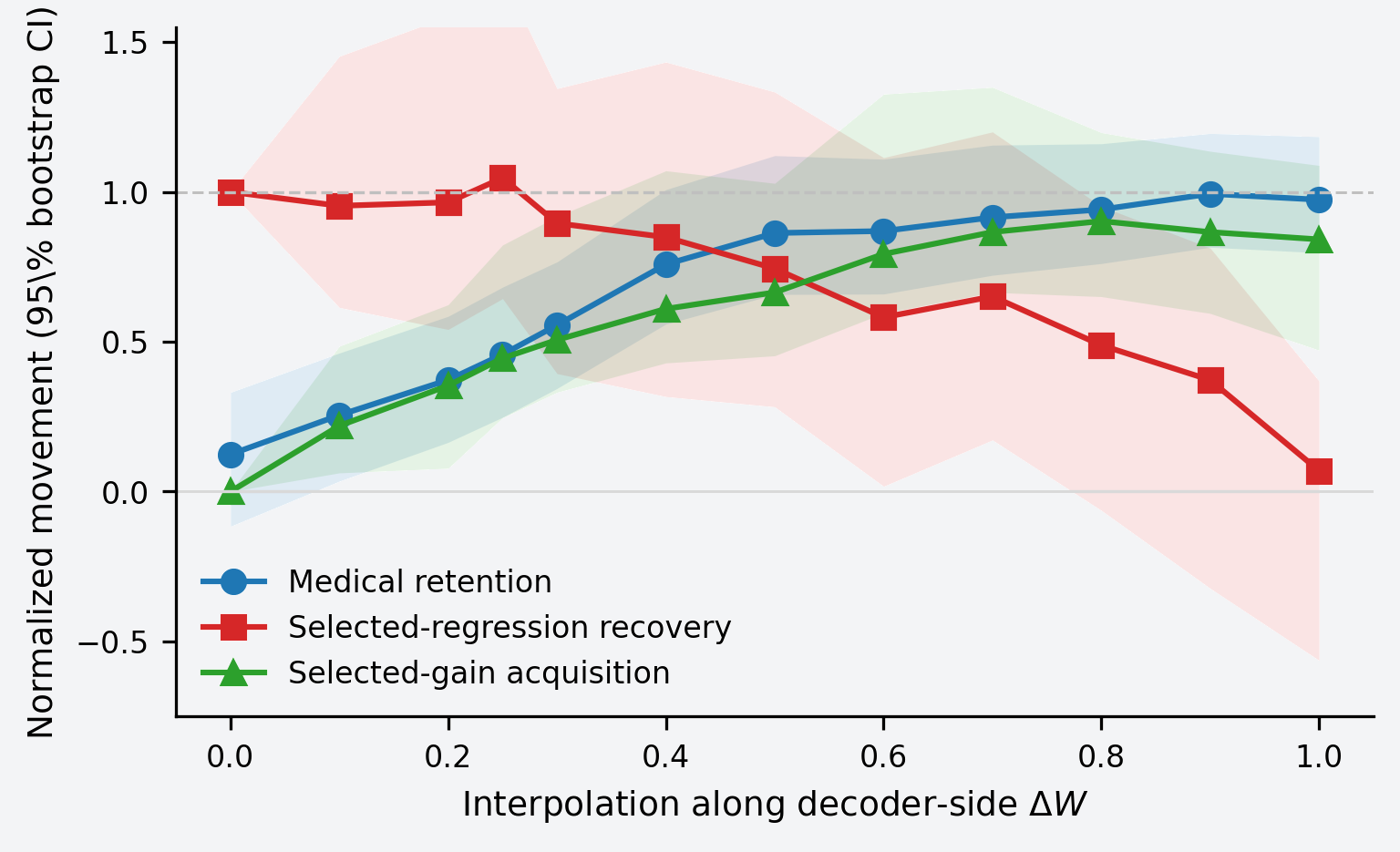}
    \caption{Bootstrap confidence bands for the Gemma--MedGemma full decoder
    path. Medical reconstruction is the most stable readout; selected-slice
    diagnostics have wider intervals and are interpreted directionally.}
    \label{fig:full-path-ci-main}
\end{figure}

\begin{figure*}[!tbp]
    \centering
    \begin{subfigure}{0.88\textwidth}
        \centering
        \includegraphics[width=\linewidth]{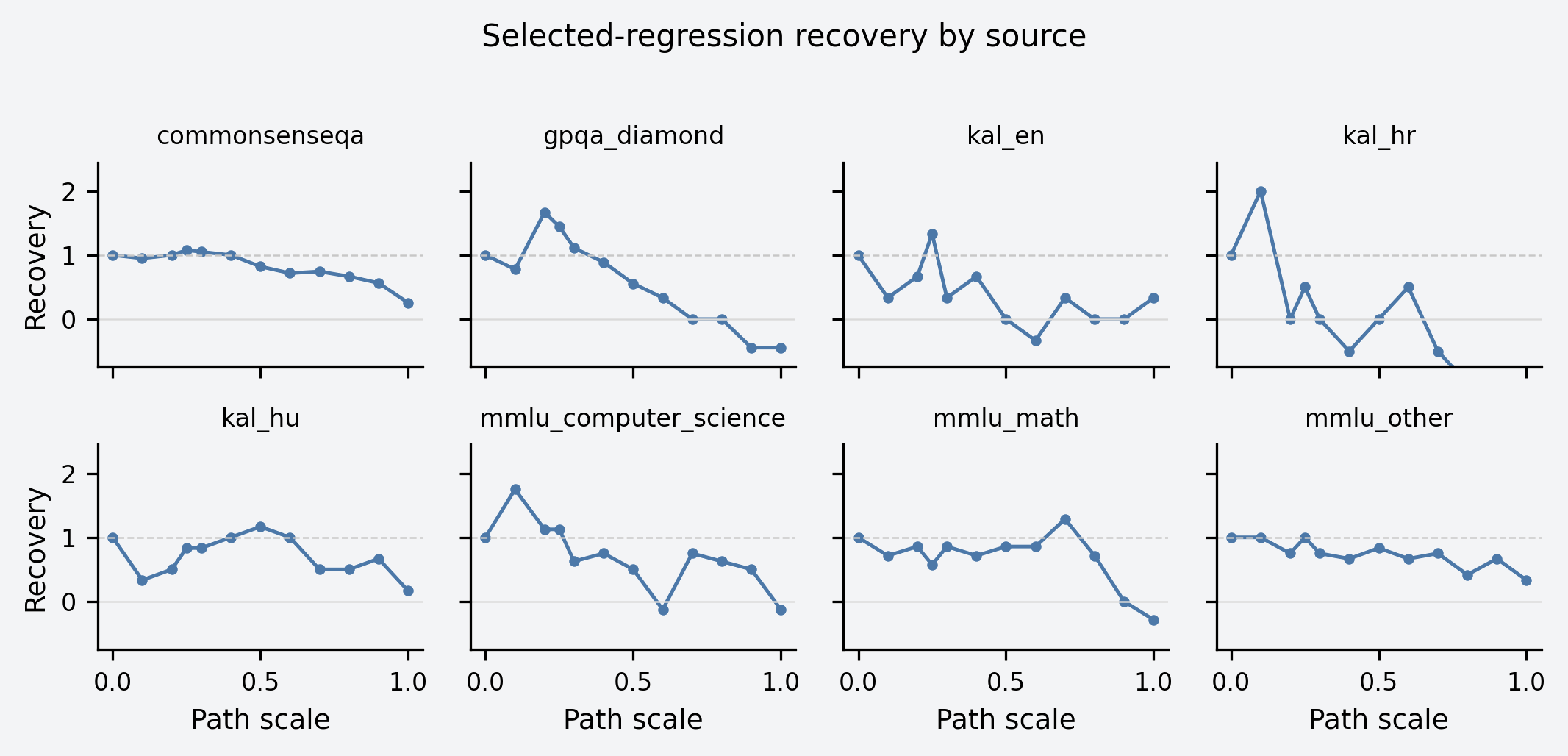}
        \caption{Selected-regression recovery}
    \end{subfigure}

    \vspace{0.2em}
    \begin{subfigure}{0.88\textwidth}
        \centering
        \includegraphics[width=\linewidth]{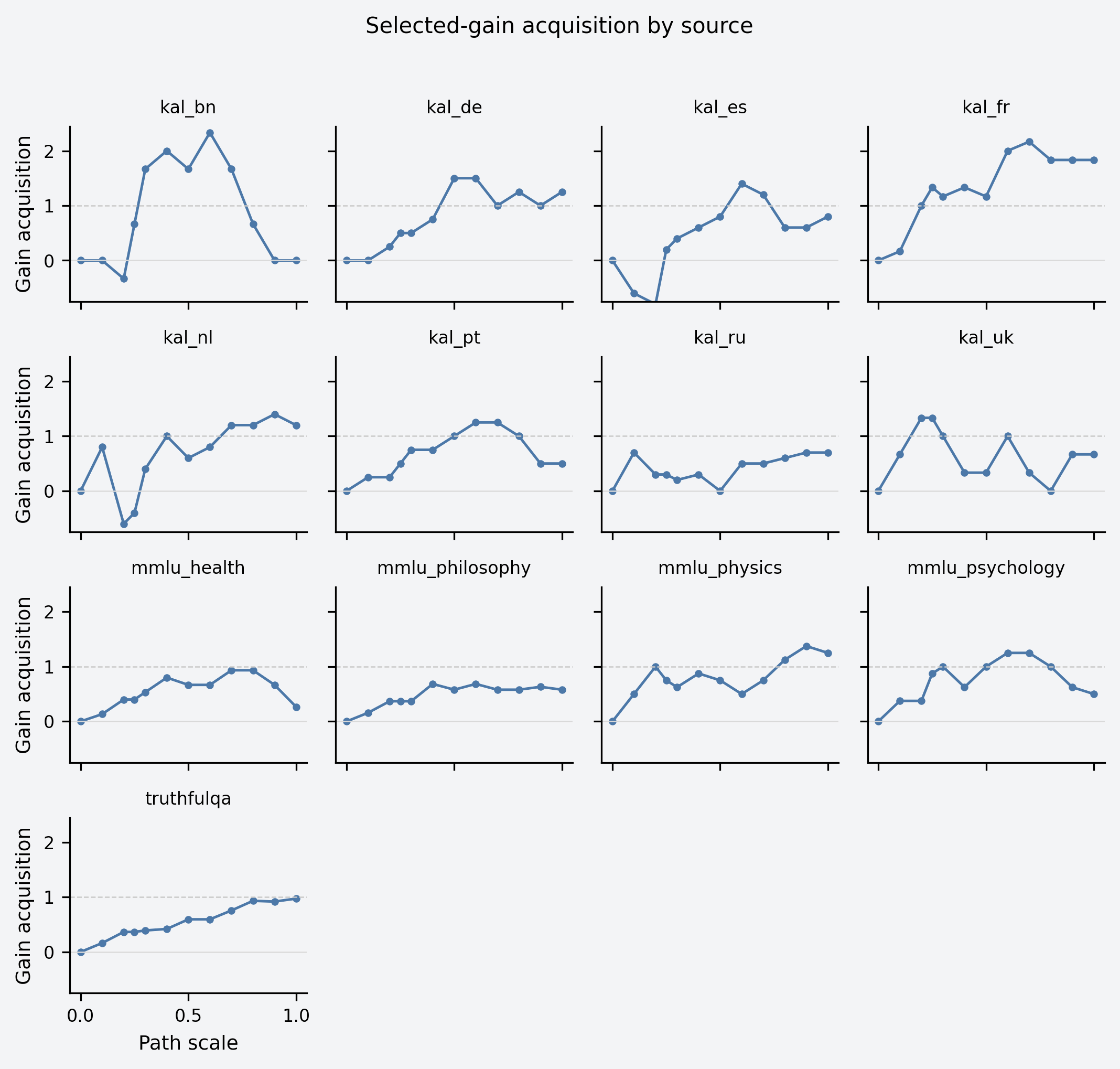}
        \caption{Selected-gain acquisition}
    \end{subfigure}
    \caption{Per-source selected-slice diagnostics along the full decoder-side
    path. These views are conditional on endpoint slice selection and are used as
    diagnostics rather than broad non-clinical estimates.}
    \label{fig:appendix-per-source-v5}
\end{figure*}

\subsection{Component Buckets and Matched Controls}

Table~\ref{tab:appendix-bucket-best-v5} reports the best observed point for each
scoped component-family path, Table~\ref{tab:appendix-matched-controls-v5}
reports 10 seeded matched controls under the same evaluation readouts, and
Table~\ref{tab:appendix-anchor-sweep-v5} reports the anchored rollback sweeps
used in Section~\ref{sec:components}. The controls are not exhaustive null
distributions, but they
make the size- and update-mass checks less dependent on a single random draw.

\begin{table}[H]
    \centering
    \scriptsize
    \setlength{\tabcolsep}{3pt}
    \begin{tabular}{@{}p{0.20\linewidth}rrrr@{}}
        \toprule
        Bucket & Matrices & Param. \% & \(\Delta W\) energy \% & Role \\
        \midrule
        Full decoder & 238 & 100.0 & 100.0 & Reference \\
        Attention family & 136 & 16.7 & 35.0 & Component \\
        MLP family & 102 & 83.3 & 65.0 & Component \\
        Attn+\texttt{down} & 170 & 44.4 & 44.2 & Mixed \\
        Full--\texttt{down} & 204 & 72.2 & 90.8 & Leave-one-out \\
        \bottomrule
    \end{tabular}
    \caption{Bucket definitions for decoder-side component-family sweeps. Shares
    are relative to the full decoder-side bucket.}
    \label{tab:appendix-bucket-definitions}
\end{table}

\begin{table}[H]
    \centering
    \scriptsize
    \setlength{\tabcolsep}{2pt}
    \resizebox{\linewidth}{!}{\begin{tabular}{lrrr}
\toprule
Bucket & Scale & Med. ret. [CI] & Reg. rec. [CI] \\
\midrule
Attention & 0.70 & 0.438 [0.205, 0.673] & 0.767 [-0.122, 1.276] \\
MLP & 0.80 & 0.797 [0.608, 1.029] & 0.628 [-0.154, 1.193] \\
Attention+down & 0.70 & 0.601 [0.374, 0.849] & 0.744 [0.020, 1.485] \\
Full-down & 0.70 & 0.667 [0.454, 0.892] & 0.465 [-0.286, 0.922] \\
\bottomrule
\end{tabular}
}
    \caption{Best observed scoped component-family points with bootstrap
    intervals.}
    \label{tab:appendix-bucket-best-v5}
\end{table}

\begin{table}[H]
    \centering
    \scriptsize
    \setlength{\tabcolsep}{2pt}
    \resizebox{\linewidth}{!}{\begin{tabular}{llrrrr}
\toprule
Target & Control & $n$ & Med. retention & Reg. recovery & Structured \\
\midrule
Attention+down & delta-energy & 10 & 0.416$\pm$0.151 & 0.431$\pm$0.136 & 0.510/0.570 \\
Attention+down & matrix-count & 10 & 0.703$\pm$0.112 & 0.388$\pm$0.159 & 0.510/0.570 \\
Attention+down & parameter-count & 10 & 0.417$\pm$0.086 & 0.585$\pm$0.166 & 0.510/0.570 \\
MLP & delta-energy & 10 & 0.563$\pm$0.100 & 0.262$\pm$0.242 & 0.699/0.279 \\
MLP & matrix-count & 10 & 0.437$\pm$0.077 & 0.513$\pm$0.224 & 0.699/0.279 \\
MLP & parameter-count & 10 & 0.814$\pm$0.067 & 0.236$\pm$0.175 & 0.699/0.279 \\
\bottomrule
\end{tabular}
}
    \caption{Matched random controls over 10 seeded draws. Entries report mean
    $\pm$ standard deviation; the final column gives the corresponding
    structured-bucket medical retention / selected-regression recovery, with
    both controls and structured comparators evaluated at \(t=1\).}
    \label{tab:appendix-matched-controls-v5}
\end{table}

\begin{figure}[H]
    \centering
    \includegraphics[width=\linewidth]{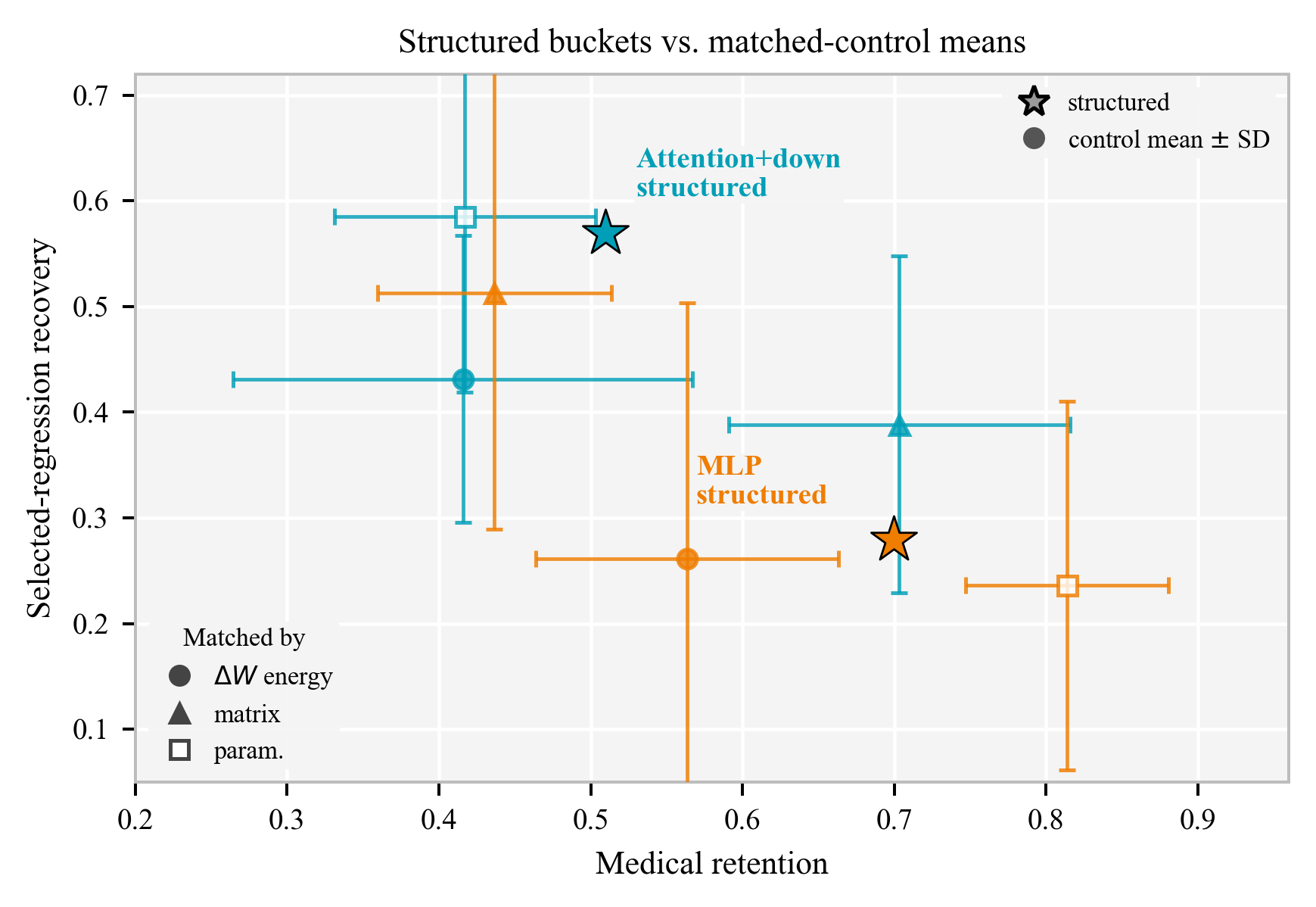}
    \caption{Matched random buckets approach the structured component buckets.
    Stars denote the structured MLP and attention+\texttt{down\_proj} buckets; other
    markers are 10-seed random-control means, with one-standard-deviation bars,
    matched by parameter count, matrix count, or \(\Delta W\) energy. Because
    several controls approach the stars, the plot supports partial sufficiency
    rather than unique component-family localization.}
    \label{fig:appendix-bucket-control-scatter}
\end{figure}

\begin{table}[H]
    \centering
    \footnotesize
    \setlength{\tabcolsep}{5pt}
    \begin{tabular}{lrrr}
\toprule
Rolled-back family & $\alpha$ & Med. ret. & Reg. rec. \\
\midrule
Attention & 0.00 & 0.699 & 0.279 \\
Attention & 0.25 & 0.784 & 0.419 \\
Attention & 0.50 & 0.902 & 0.314 \\
Attention & 0.75 & 0.993 & 0.256 \\
Attention & 1.00 & 0.974 & 0.070 \\
\midrule
MLP & 0.00 & 0.359 & 0.593 \\
MLP & 0.25 & 0.699 & 0.477 \\
MLP & 0.50 & 0.876 & 0.384 \\
MLP & 0.75 & 0.980 & 0.302 \\
MLP & 1.00 & 0.974 & 0.070 \\
\bottomrule
\end{tabular}

    \caption{Anchored rollback paths. Lower \(\alpha\) rolls back the named
    family while leaving the rest of the decoder update fixed at the specialist
    endpoint.}
    \label{tab:appendix-anchor-sweep-v5}
\end{table}

\subsection{Qwen2.5--HuatuoGPT-o1 Replication}
\label{sec:appendix-qwen}

The replication uses the same 1,810 medical and 7,325 non-clinical examples,
scoring rule, normalized readouts, and bootstrap procedure. Its
endpoint-conditioned regression and gain slices are selected within the pair.
The 196 aligned decoder projection matrices contain approximately 6.53B
non-embedding parameters.

\begin{table}[H]
    \centering
    \scriptsize
    \begin{tabular}{@{}lrrr@{}}
\toprule
Endpoint view & Estimate & \multicolumn{2}{c}{95\% CI} \\
\midrule
Medical composite & +0.0453 & +0.0254 & +0.0641 \\
Selected regressions & -0.0617 & -0.0813 & -0.0417 \\
Selected gains & +0.0554 & +0.0387 & +0.0721 \\
\bottomrule
\end{tabular}

\vspace{0.8em}

\begin{tabular}{@{}rrrrr@{}}
\toprule
$t$ & Med. acc. & Med. ret. & Reg. rec. & Gain acq. \\
\midrule
0.00 & 0.6481 & 0.000 & 1.000 & 0.000 \\
0.50 & 0.6873 & 0.866 & 0.789 & 0.824 \\
0.60 & 0.6961 & 1.061 & 0.585 & 0.919 \\
0.80 & 0.6972 & 1.085 & 0.458 & 1.103 \\
1.00 & 0.7017 & 1.183 & 0.556 & 1.147 \\
\bottomrule
\end{tabular}

    \caption{Qwen--HuatuoGPT endpoint movements and evaluated full-path grid.}
    \label{tab:appendix-qwen-path}
\end{table}

\begin{table}[H]
    \centering
    \scriptsize
    \resizebox{\linewidth}{!}{\begin{tabular}{@{}lrrrr@{}}
\toprule
Group & Matrices & Param. share & $\Delta W$ energy & Mean rank \\
\midrule
Attention & 112 & 12.6\% & 12.2\% & 477.7 \\
MLP & 84 & 87.4\% & 87.8\% & 1522.3 \\
\quad\texttt{gate\_proj} & 28 & 29.1\% & 30.6\% & 1471 \\
\quad\texttt{up\_proj} & 28 & 29.1\% & 29.0\% & 1354 \\
\quad\texttt{down\_proj} & 28 & 29.1\% & 28.2\% & 1742 \\
\bottomrule
\end{tabular}
}
    \caption{Qwen decoder projection coverage and update geometry.}
    \label{tab:appendix-qwen-coverage}
\end{table}

\begin{figure}[H]
    \centering
    \includegraphics[width=\linewidth]{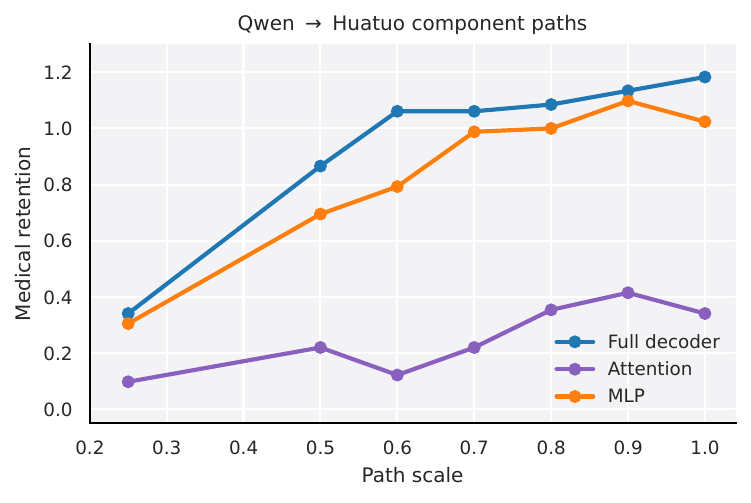}
    \caption{Qwen structured-family medical-retention trajectories at the
    evaluated path points. MLP is strongest, but its size motivates matched
    controls.}
    \label{fig:appendix-qwen-families}
\end{figure}

\begin{table*}[!tbp]
    \centering
    \scriptsize
    \begin{tabular}{@{}llrrrr@{}}
\toprule
Target & Match & Med. mean & Range & Reg. mean & Gain mean \\
\midrule
Attn+down & $\Delta W$ energy & 0.660 & 0.573--0.780 & 0.744 & 0.674 \\
Attn+down & Matrices & 1.117 & 1.061--1.256 & 0.685 & 0.941 \\
Attn+down & Parameters & 0.667 & 0.476--0.817 & 0.799 & 0.707 \\
MLP & $\Delta W$ energy & 1.098 & 1.037--1.159 & 0.548 & 1.102 \\
MLP & Matrices & 0.680 & 0.427--0.902 & 0.837 & 0.713 \\
MLP & Parameters & 1.115 & 1.024--1.195 & 0.527 & 1.082 \\
\bottomrule
\end{tabular}

    \caption{Qwen matched controls at \(t=1\) over 10 seeded draws. Ranges show
    the observed minimum and maximum, not standard deviations or confidence
    intervals.}
    \label{tab:appendix-qwen-controls}
\end{table*}

\begin{figure}[H]
    \centering
    \includegraphics[width=\linewidth]{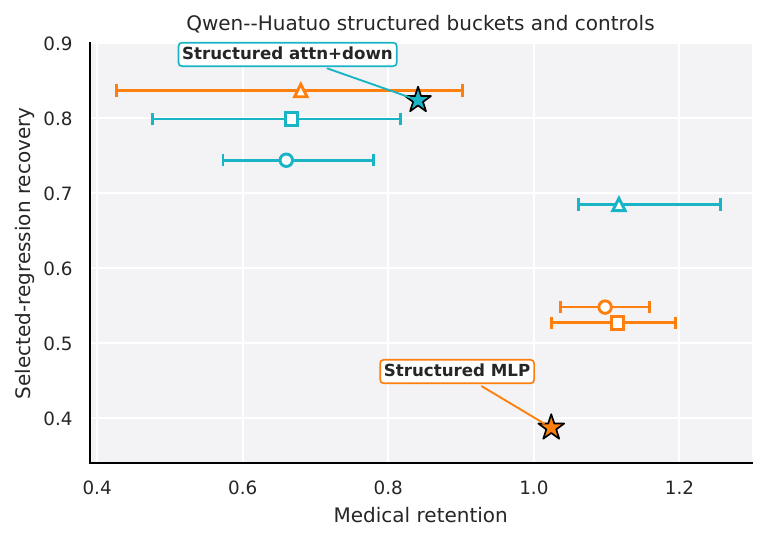}
    \caption{Qwen control means and observed ranges over 10 seeded draws. Stars mark the
    corresponding structured bucket at \(t=1\).}
    \label{fig:appendix-qwen-controls}
\end{figure}

\begin{table*}[!tbp]
    \centering
    \scriptsize
    \begin{tabular}{@{}lr@{}}
\toprule
Module-only path & Med. retention \\
\midrule
\texttt{down\_proj} & 0.512 \\
\texttt{up\_proj} & 0.439 \\
\texttt{gate\_proj} & 0.366 \\
\texttt{o\_proj} & 0.183 \\
\texttt{v\_proj} & 0.061 \\
\texttt{q\_proj} & 0.012 \\
\texttt{k\_proj} & -0.037 \\
\bottomrule
\end{tabular}
\qquad
\begin{tabular}{@{}lr@{}}
\toprule
Leave-one-module-out path & Med. retention \\
\midrule
Full minus \texttt{up\_proj} & 1.183 \\
Full minus \texttt{down\_proj} & 1.159 \\
Full minus \texttt{q\_proj} & 1.146 \\
Full minus \texttt{gate\_proj} & 1.146 \\
Full minus \texttt{v\_proj} & 1.110 \\
Full minus \texttt{k\_proj} & 1.098 \\
Full minus \texttt{o\_proj} & 1.085 \\
\bottomrule
\end{tabular}

    \caption{Qwen module-only and leave-one-module-out paths at \(t=1\).}
    \label{tab:appendix-qwen-modules}
\end{table*}

\begin{table*}[!tbp]
    \centering
    \scriptsize
    \resizebox{\textwidth}{!}{\begin{tabular}{@{}lccc@{}}
\toprule
Full rollback & Med. retention & Reg. recovery & Gain acquisition \\
\midrule
Attention & 0.756 [0.494, 1.000] & 0.387 [0.212, 0.600] & 0.882 [0.723, 1.084] \\
MLP & 0.293 [0.011, 0.541] & 0.986 [0.800, 1.234] & 0.471 [0.297, 0.677] \\
Attn+down & 0.549 [0.295, 0.841] & 0.570 [0.342, 0.789] & 1.000 [0.790, 1.296] \\
Full-minus-down & 0.317 [0.085, 0.557] & 0.866 [0.671, 1.083] & 0.559 [0.394, 0.774] \\
\bottomrule
\end{tabular}
}
    \caption{Qwen full rollback results with 95\% bootstrap intervals.}
    \label{tab:appendix-qwen-rollback}
\end{table*}

\begin{figure}[H]
    \centering
    \includegraphics[width=\linewidth]{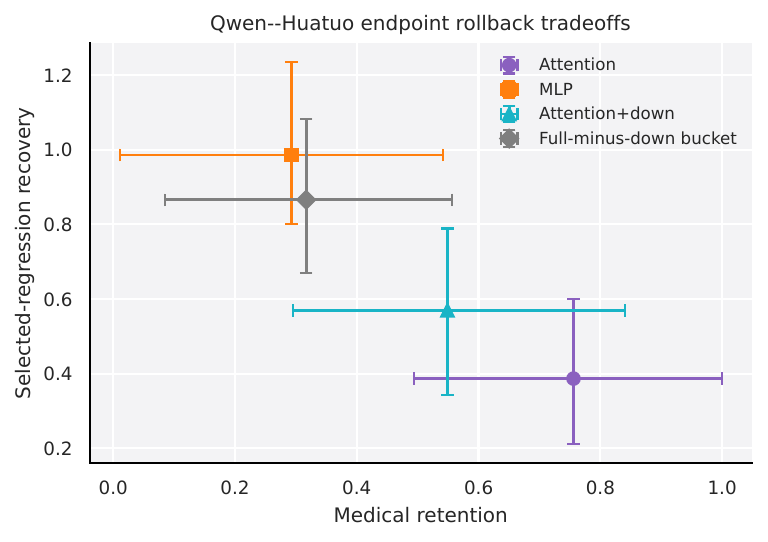}
    \caption{Qwen rollback tradeoff. No tested family simultaneously preserves
    the medical movement and fully recovers selected regressions.}
    \label{fig:appendix-qwen-rollback}
\end{figure}

\subsection{MLP Neuron-Level Activation Diagnostic}
\label{sec:appendix-neuron-diagnostic}

As a secondary diagnostic, we ask whether the coarse component results hide a
simpler pattern at the level of MLP intermediate neurons. We sample 64 examples
from each of the medical, selected-regression, and selected-gain views and run
both checkpoints on the same prompts. For each decoder layer, we capture the
last-token MLP intermediate state
\(\mathrm{act}(\mathrm{gate\_proj}(x)) \odot \mathrm{up\_proj}(x)\), yielding
arrays of shape \(64 \times 34 \times 10240\) for each diagnostic group and
checkpoint. We define drift as the mean absolute Gemma--MedGemma activation
difference, and define a proxy importance score as the MedGemma activation
magnitude multiplied by the corresponding \(\mathrm{down\_proj}\) column norm.
These are descriptive proxies, not causal neuron ablations.

Table~\ref{tab:appendix-neuron-overlap-v5} reports top-\(k\) Jaccard overlaps
between the resulting neuron rankings. The main pattern is conservative:
high-importance neurons are strongly shared across medical, selected-regression,
and selected-gain groups, whereas importance and specialization drift overlap
only partially. Drift rankings are more similar across medical and regression
groups than either drift ranking is to medical importance. Layer summaries show
the same separation: mean drift is strongest in late layers, while proxy
importance includes layer 33 but also earlier and middle layers. Thus this
finer-grained diagnostic does not overturn the main localization result; it
supports the view that the specialization update is structured but not cleanly
explained by a small medical-specific neuron set.

\begin{table}[H]
    \centering
    \scriptsize
    \setlength{\tabcolsep}{4pt}
    \begin{tabular}{lrrr}
        \toprule
        Ranking comparison & Top-100 & Top-1k & Top-5k \\
        \midrule
        Medical imp. vs medical drift & 0.212 & 0.255 & 0.275 \\
        Medical imp. vs regression drift & 0.250 & 0.277 & 0.283 \\
        Medical imp. vs regression imp. & 0.786 & 0.619 & 0.603 \\
        Medical drift vs regression drift & 0.515 & 0.514 & 0.520 \\
        Regression imp. vs regression drift & 0.266 & 0.278 & 0.317 \\
        Medical imp. vs gain imp. & 0.802 & 0.630 & 0.618 \\
        Regression imp. vs gain imp. & 0.852 & 0.812 & 0.797 \\
        \bottomrule
    \end{tabular}
    \caption{Top-\(k\) Jaccard overlaps for the MLP neuron diagnostic. ``Imp.''
    denotes the down-projection-weighted activation-magnitude proxy. The
    diagnostic uses 64 examples per group and is reported only as descriptive
    support for the coarse localization analysis.}
    \label{tab:appendix-neuron-overlap-v5}
\end{table}

\subsection{Additional Descriptive \texorpdfstring{\(\Delta W\)}{Delta W} Tables and Figures}
\label{sec:appendix-dw}

The main text includes the decoder-side relative-change heatmap and layerwise
RMS curve. Here we report additional descriptive \(\Delta W\) diagnostics used
to design the bucket set, including spectral energy and effective-rank views.
These are not separate intervention results.

\begin{table}[H]
    \centering
    \scriptsize
    \setlength{\tabcolsep}{3pt}
    \resizebox{\linewidth}{!}{%
    \begin{tabular}{llrrrrrr}
        \toprule
        Module & Family & All \(n\) & All rel. & All top-32 & All eff. rank & Dec. rel. & Dec. top-32 \\
        \midrule
        \texttt{q\_proj} & attention & 61 & 2.486 & 0.178 & 786.8 & 0.368 & 0.137 \\
        \texttt{v\_proj} & attention & 61 & 2.121 & 0.136 & 668.4 & 0.338 & 0.104 \\
        \texttt{k\_proj} & attention & 61 & 1.941 & 0.174 & 611.8 & 0.373 & 0.142 \\
        \texttt{down\_proj} & MLP & 34 & 0.370 & 0.039 & 2171.0 & 0.370 & 0.039 \\
        \texttt{o\_proj} & attention & 34 & 0.337 & 0.122 & 1086.6 & 0.337 & 0.122 \\
        \texttt{gate\_proj} & MLP & 34 & 0.261 & 0.033 & 2175.5 & 0.261 & 0.033 \\
        \texttt{up\_proj} & MLP & 34 & 0.243 & 0.034 & 2168.4 & 0.243 & 0.034 \\
        \bottomrule
    \end{tabular}
    }
    \caption{Module-level \(\Delta W\) summaries. ``All'' aggregates all
    selected aligned matrices, including the vision tower where applicable.
    ``Dec.'' recomputes the same relative-change and mean top-32 energy
    summaries over decoder-side matrices only. Relative change is computed as
    the aggregate Frobenius norm of \(\Delta W\) divided by the corresponding
    aggregate base-model norm.}
    \label{tab:appendix-dw-module-summary}
\end{table}

\begin{table}[H]
    \centering
    \scriptsize
    \setlength{\tabcolsep}{3pt}
    \resizebox{\linewidth}{!}{%
    \begin{tabular}{rrrrrrrl}
        \toprule
        Layer & Rel. change & RMS & Attn rel. & MLP rel. & Attn/MLP & Eff. rank & Max module \\
        \midrule
        19 & 1.286 & 0.0222 & 2.649 & 0.248 & 5.81 & 1208.2 & \texttt{v\_proj} \\
        26 & 1.283 & 0.0215 & 2.463 & 0.247 & 5.97 & 1173.2 & \texttt{v\_proj} \\
        2  & 1.254 & 0.0219 & 2.375 & 0.255 & 5.67 & 1049.2 & \texttt{q\_proj} \\
        20 & 1.241 & 0.0216 & 2.193 & 0.262 & 5.62 & 1196.8 & \texttt{v\_proj} \\
        18 & 1.239 & 0.0219 & 2.510 & 0.248 & 5.62 & 1212.1 & \texttt{v\_proj} \\
        17 & 1.239 & 0.0214 & 2.438 & 0.256 & 5.50 & 1160.1 & \texttt{v\_proj} \\
        21 & 1.203 & 0.0215 & 2.215 & 0.251 & 5.57 & 1225.7 & \texttt{v\_proj} \\
        22 & 1.198 & 0.0212 & 2.311 & 0.240 & 5.70 & 1235.9 & \texttt{v\_proj} \\
        \bottomrule
    \end{tabular}
    }
    \caption{Top decoder-side layers by layer-level relative change. The
    attention/MLP ratio compares Frobenius norms of the attention and
    feed-forward parts of each layer's \(\Delta W\).}
    \label{tab:appendix-dw-layer-summary}
\end{table}

\begin{table}[H]
    \centering
    \scriptsize
    \setlength{\tabcolsep}{3pt}
    \resizebox{\linewidth}{!}{%
    \begin{tabular}{rlrrrrr}
        \toprule
        Layer & Module & Rel. change & RMS & Top-32 energy & Eff. rank & \(k_{80}\) \\
        \midrule
        27 & \texttt{k\_proj} & 0.854 & 0.0173 & 0.165 & 665.4 & 476 \\
        7  & \texttt{k\_proj} & 0.843 & 0.0176 & 0.152 & 673.8 & 455 \\
        27 & \texttt{v\_proj} & 0.837 & 0.0170 & 0.084 & 822.9 & 548 \\
        11 & \texttt{down\_proj} & 0.830 & 0.0044 & 0.053 & 2070.3 & 1441 \\
        19 & \texttt{o\_proj} & 0.825 & 0.0083 & 0.114 & 1073.8 & 734 \\
        7  & \texttt{v\_proj} & 0.814 & 0.0170 & 0.099 & 771.4 & 508 \\
        15 & \texttt{down\_proj} & 0.797 & 0.0042 & 0.046 & 2119.2 & 1461 \\
        8  & \texttt{k\_proj} & 0.767 & 0.0167 & 0.180 & 632.7 & 439 \\
        18 & \texttt{o\_proj} & 0.765 & 0.0083 & 0.128 & 1052.7 & 738 \\
        8  & \texttt{v\_proj} & 0.729 & 0.0159 & 0.093 & 790.1 & 522 \\
        \bottomrule
    \end{tabular}
    }
    \caption{Top decoder-side matrices by matrix-level relative change.
    \(k_{80}\) is the number of singular directions needed to explain 80\% of
    the matrix \(\Delta W\) energy.}
    \label{tab:appendix-dw-matrix-summary}
\end{table}

\begin{figure}[H]
    \centering
    \begin{subfigure}{\linewidth}
        \centering
        \includegraphics[width=\linewidth]{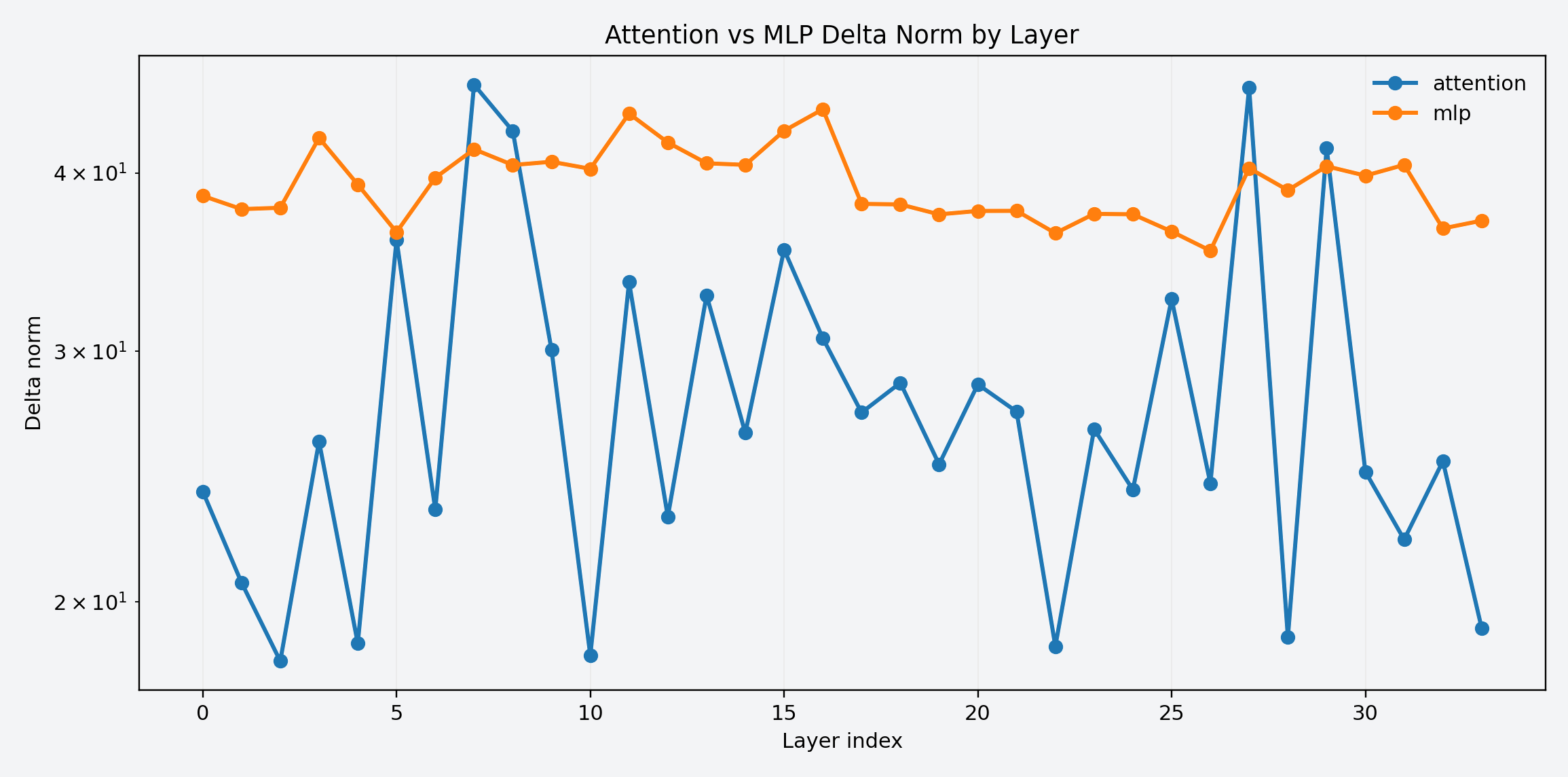}
        \caption{Attention vs MLP delta by layer}
    \end{subfigure}
    \vspace{0.25em}

    \begin{subfigure}{\linewidth}
        \centering
        \includegraphics[width=\linewidth]{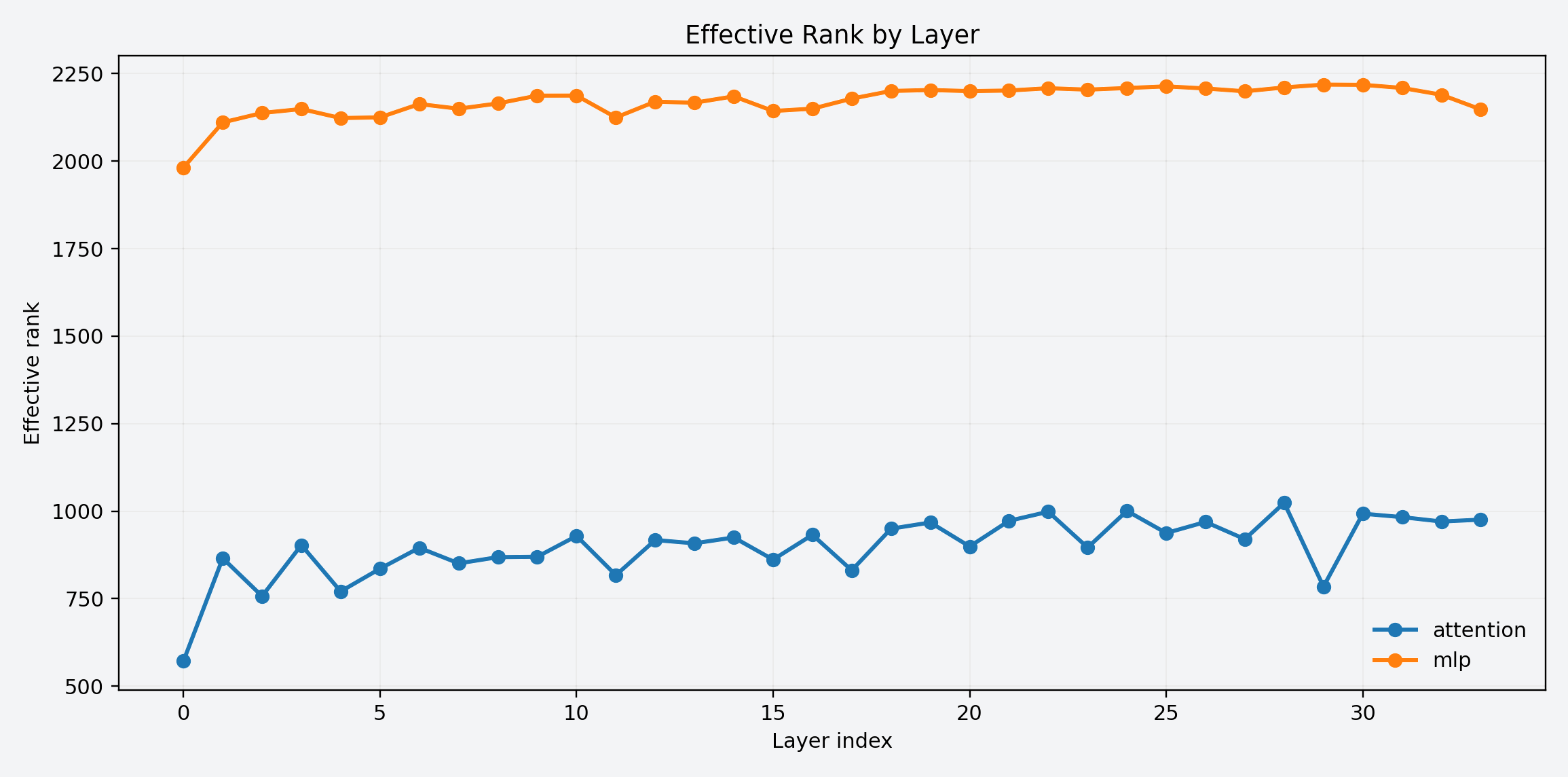}
        \caption{Effective rank by layer}
    \end{subfigure}
    \caption{Additional decoder-side layer diagnostics. These views show that
    the decoder-side update is depth-structured, while the intervention
    analysis in the main text focuses on component-level bucket sufficiency.}
    \label{fig:appendix-language-layer-dw}
\end{figure}

\begin{figure*}[!tbp]
    \centering
    \begin{subfigure}{0.49\textwidth}
        \centering
        \includegraphics[width=\linewidth]{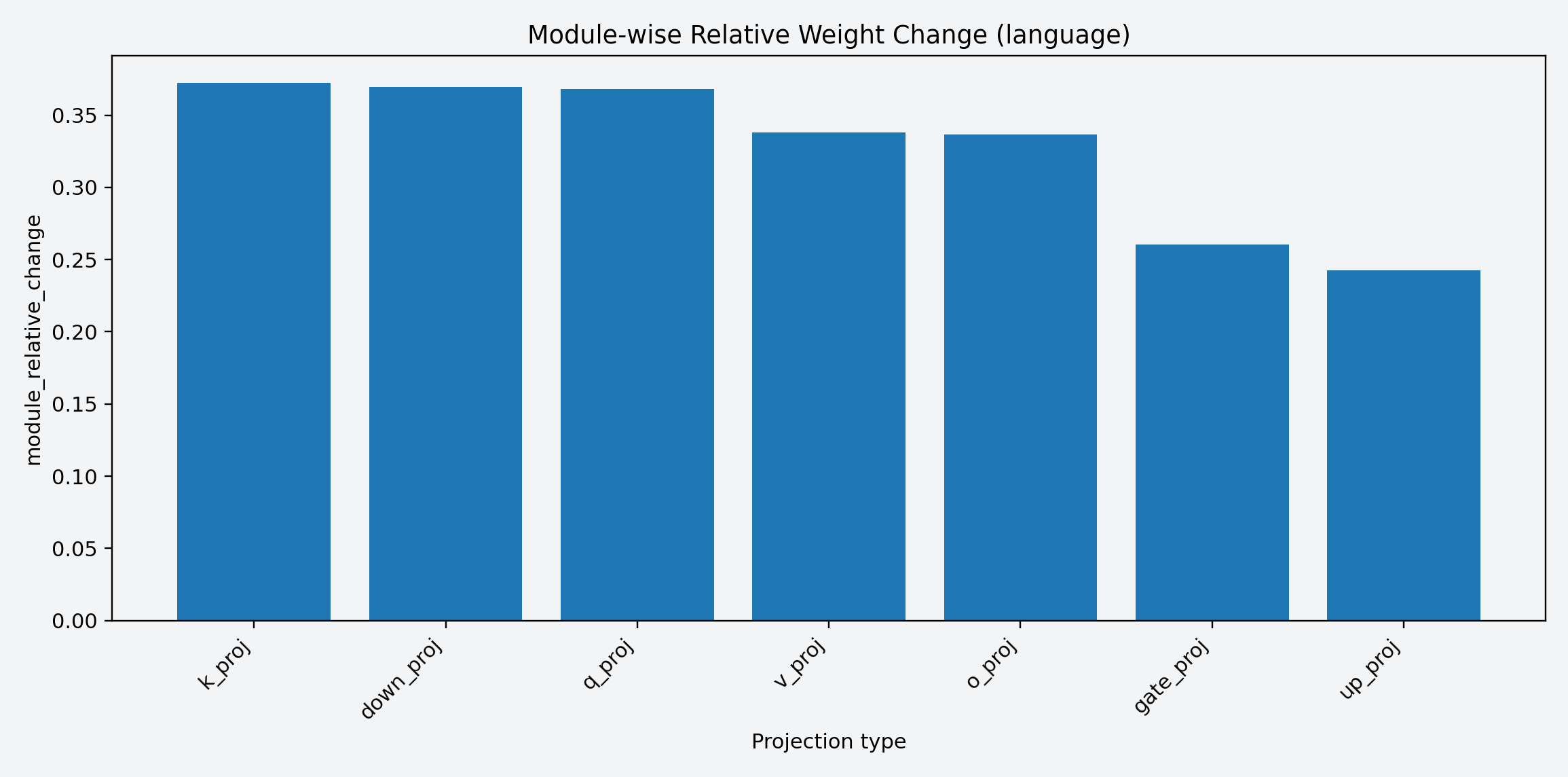}
        \caption{Module relative change}
    \end{subfigure}
    \hfill
    \begin{subfigure}{0.49\textwidth}
        \centering
        \includegraphics[width=\linewidth]{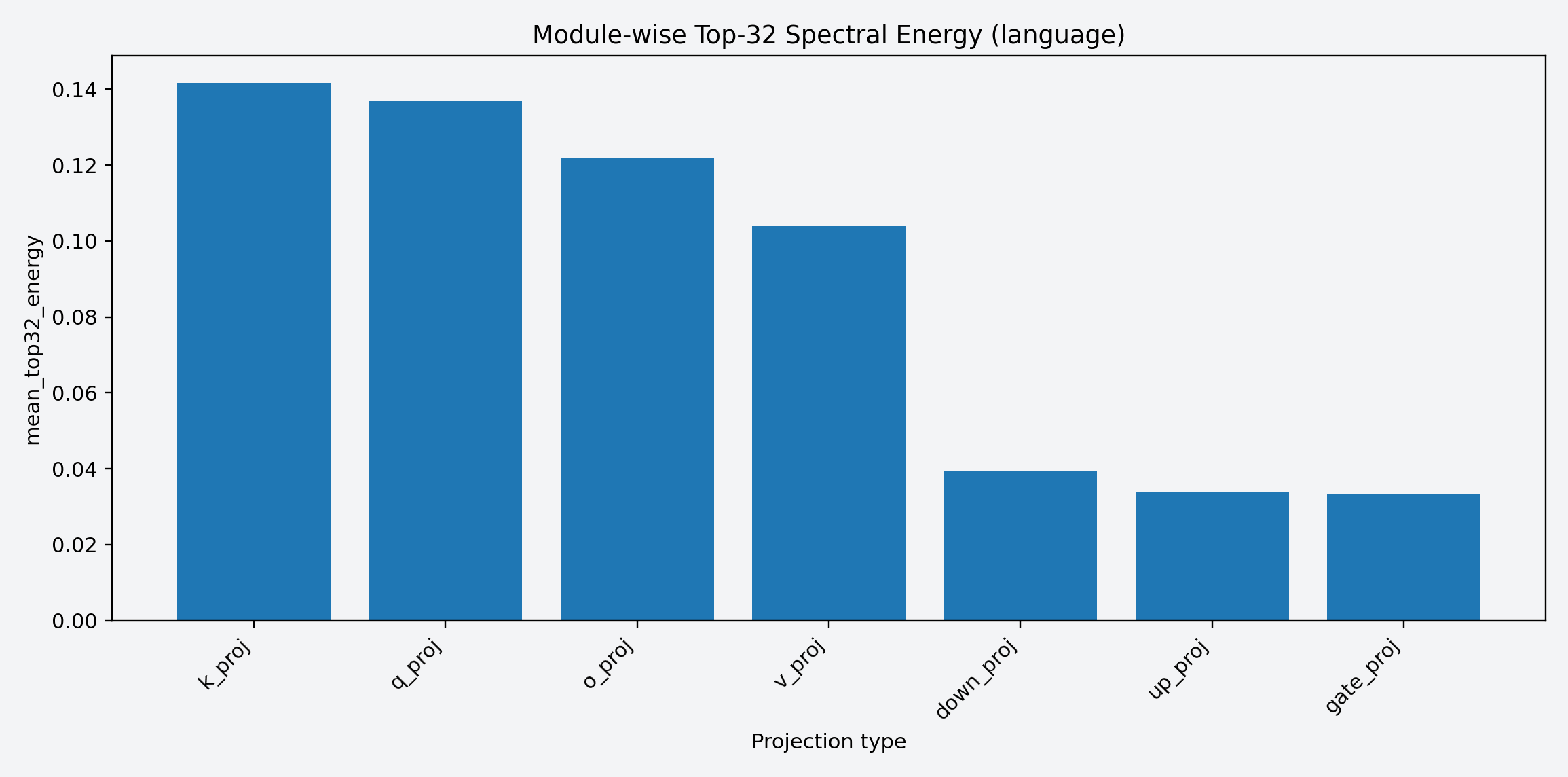}
        \caption{Module top-32 energy}
    \end{subfigure}
    \caption{Decoder-side module diagnostics complementing
    Table~\ref{tab:appendix-dw-module-summary}.}
    \label{fig:appendix-language-module-dw}
\end{figure*}

\begin{figure*}[!tbp]
    \centering
    \begin{subfigure}{0.49\textwidth}
        \centering
        \includegraphics[width=\linewidth]{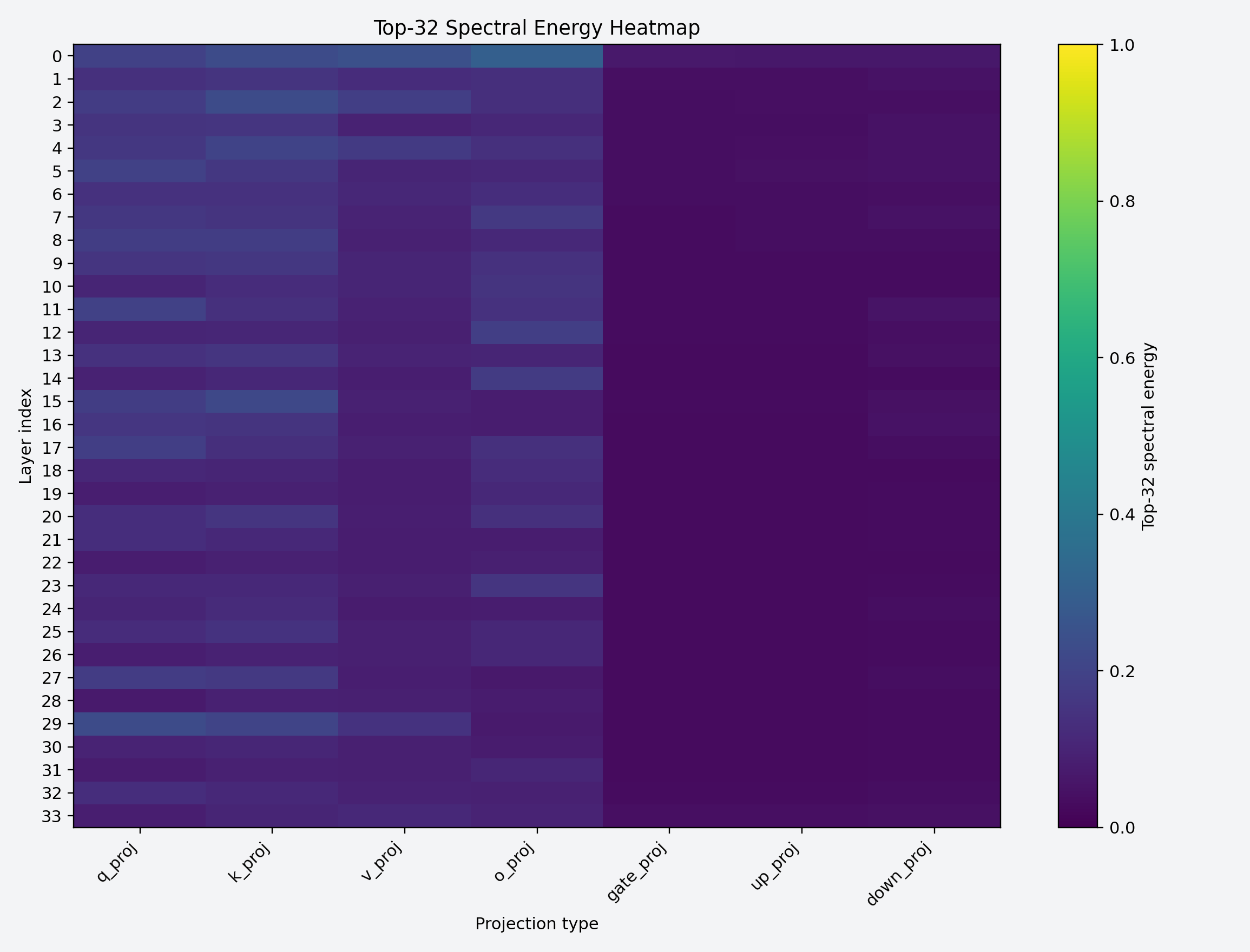}
        \caption{Top-32 energy heatmap}
    \end{subfigure}
    \hfill
    \begin{subfigure}{0.49\textwidth}
        \centering
        \includegraphics[width=\linewidth]{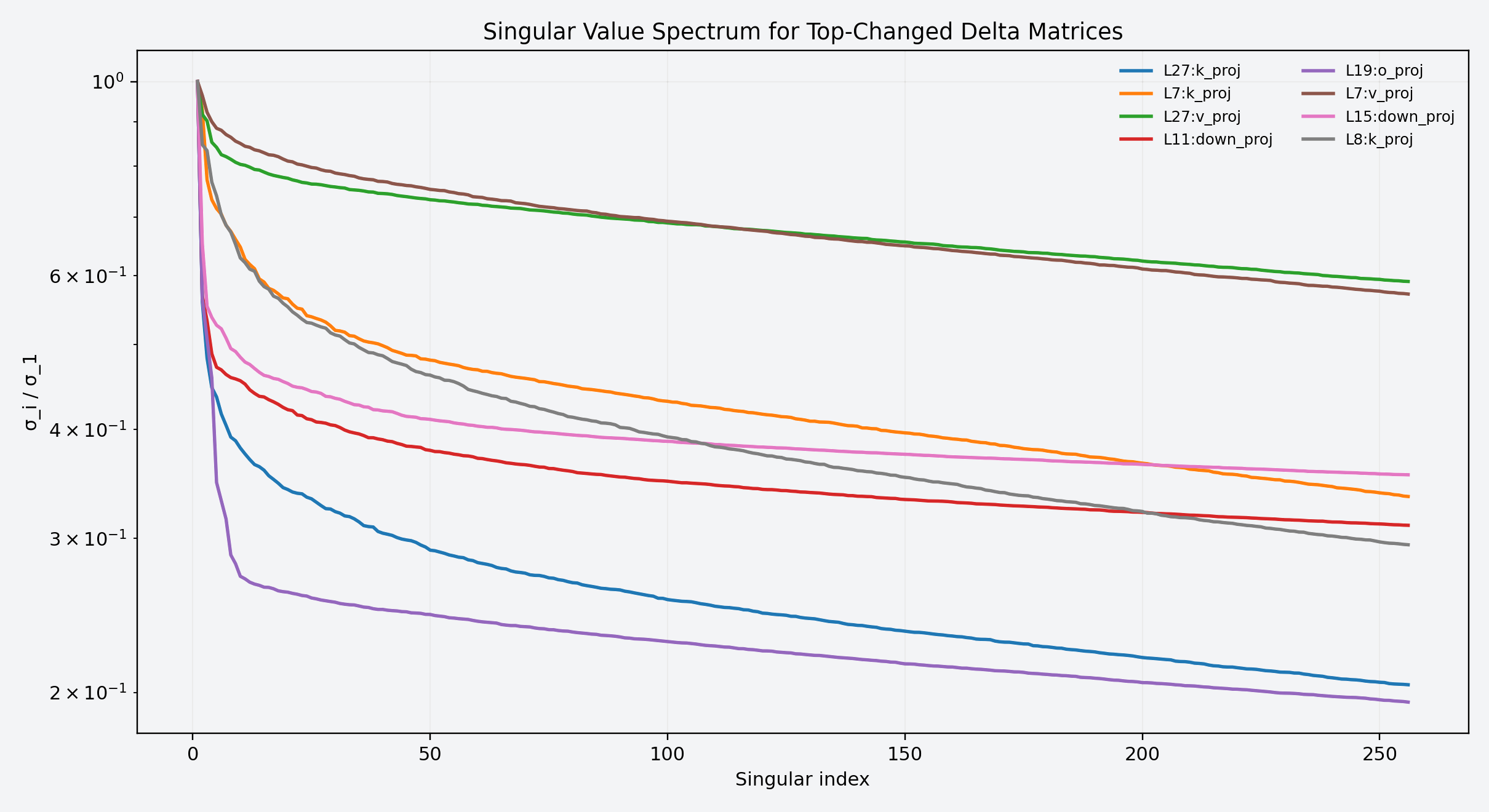}
        \caption{Singular-value spectra}
    \end{subfigure}
    \caption{Additional spectral diagnostics for decoder-side
    \(\Delta W\) matrices.}
    \label{fig:appendix-language-spectral-dw}
\end{figure*}

\begin{figure*}[!tbp]
    \centering
    \begin{subfigure}{0.49\textwidth}
        \centering
        \includegraphics[width=\linewidth]{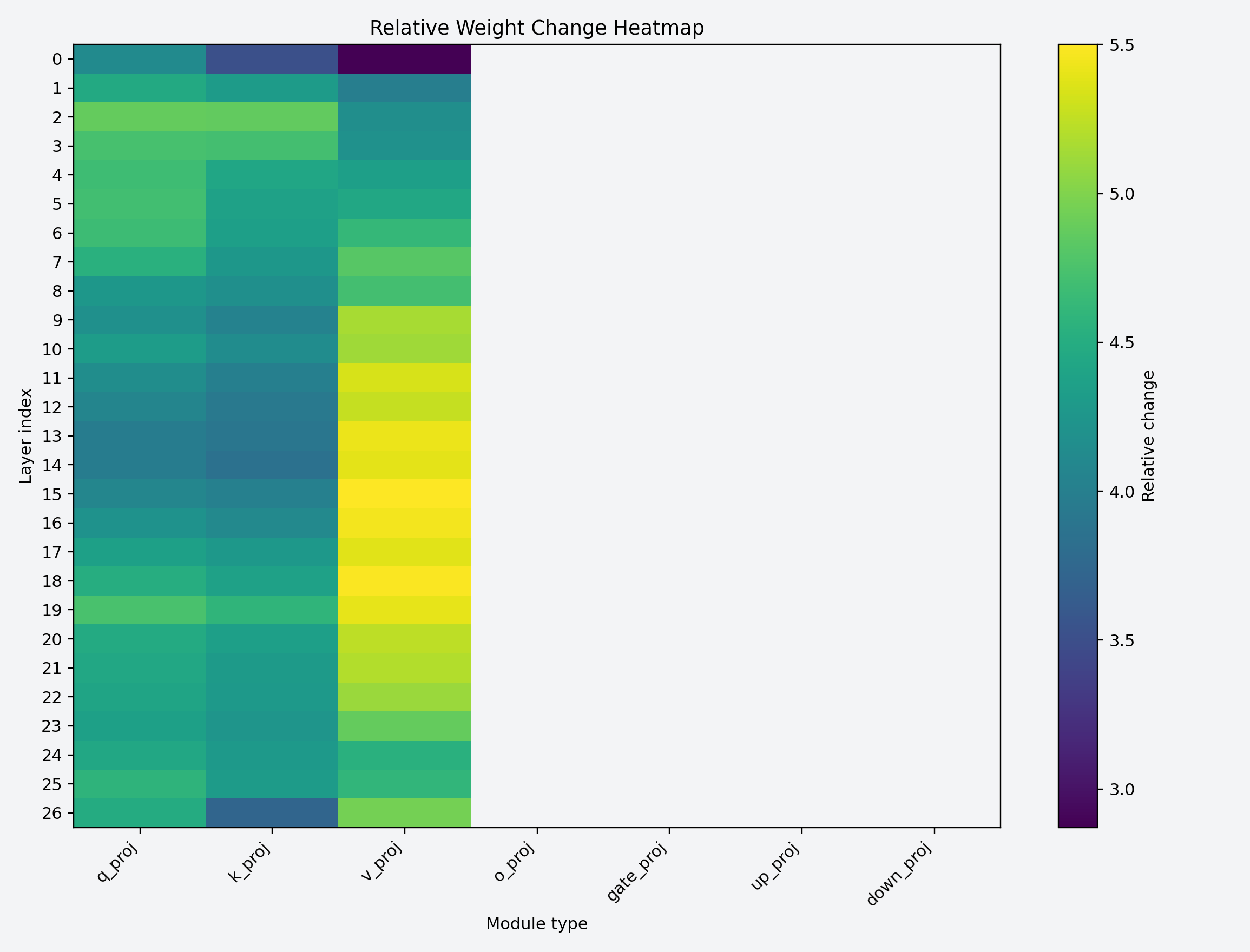}
        \caption{Vision relative-change heatmap}
    \end{subfigure}
    \hfill
    \begin{subfigure}{0.49\textwidth}
        \centering
        \includegraphics[width=\linewidth]{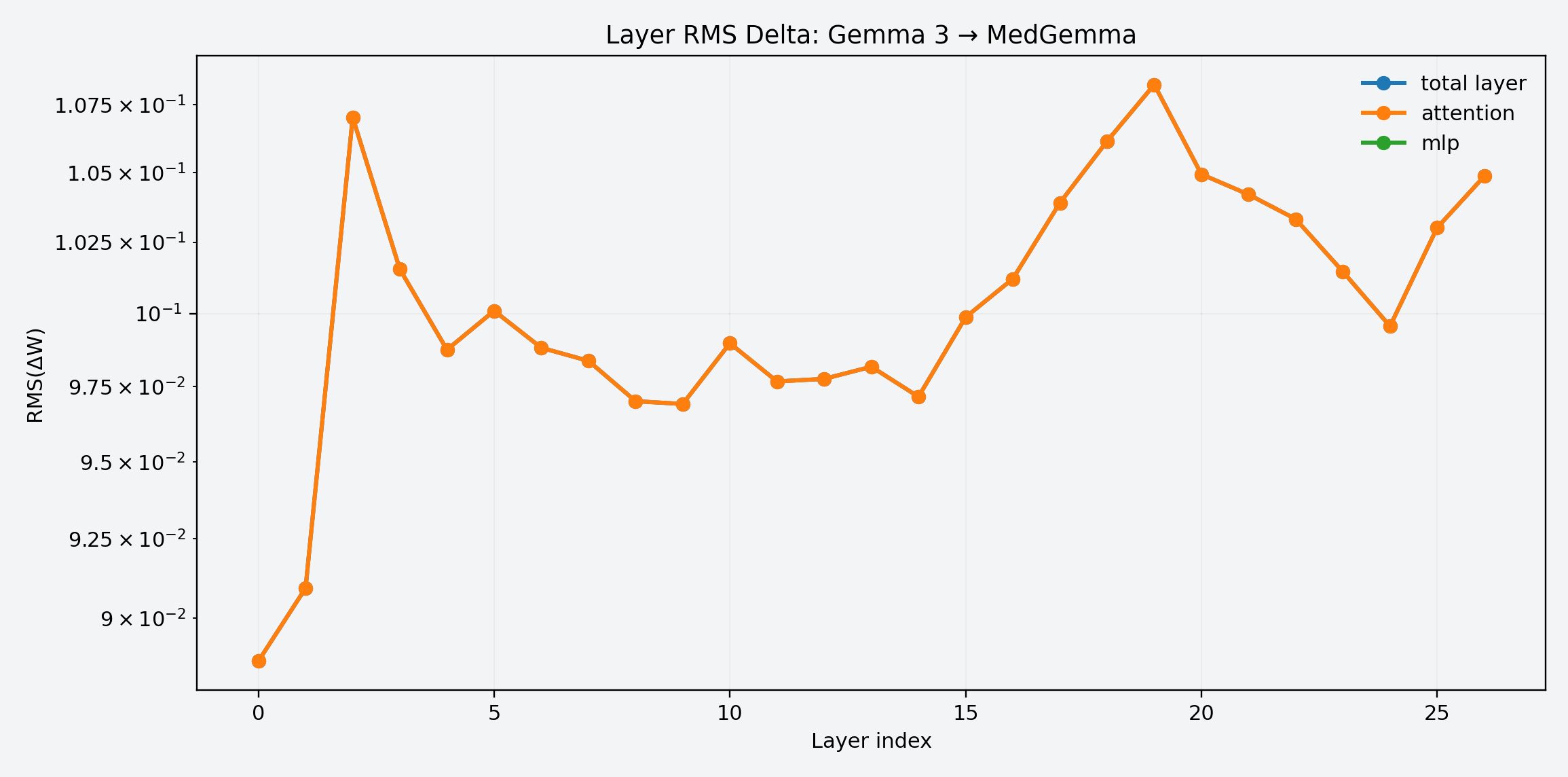}
        \caption{Vision layer RMS delta}
    \end{subfigure}
    \caption{Vision-side descriptive \(\Delta W\) context. These plots explain
    why the global raw \(\Delta W\) screen contains large vision-tower changes,
    even though the main intervention analysis targets the decoder backbone
    for text-only medical evaluation.}
    \label{fig:appendix-vision-dw}
\end{figure*}

\end{document}